\documentclass{article}

\usepackage{microtype}
\usepackage{graphicx}
\usepackage{subcaption}
\usepackage{booktabs} % for professional tables
\usepackage[colorlinks=true, linkcolor=blue, citecolor=blue]{hyperref}

 \usepackage[preprint]{icml2026}

\usepackage{amsmath}
\usepackage{amssymb}
\usepackage{mathtools}
\usepackage{amsthm}
\usepackage[table]{xcolor}

\usepackage[capitalize,noabbrev]{cleveref}
\usepackage{tikz}
\usepackage{subcaption}
\usepackage{graphicx}
\usepackage{amsmath,amsfonts,bm}
\usepackage{booktabs}
\usepackage{pgfplots}
\def\eqref#1{equation~\ref{#1}}
\def\1{\bm{1}}

\DeclareMathAlphabet{\mathsfit}{\encodingdefault}{\sfdefault}{m}{sl}
\SetMathAlphabet{\mathsfit}{bold}{\encodingdefault}{\sfdefault}{bx}{n}

\def\gA{{\mathcal{A}}}

\def\gC{{\mathcal{C}}}

\def\gL{{\mathcal{L}}}

\def\gS{{\mathcal{S}}}

\newcommand{\E}{\mathbb{E}}

\theoremstyle{plain}

\theoremstyle{definition}

\theoremstyle{remark}

\usepackage[textsize=tiny]{todonotes}
\icmltitlerunning{Optimistic World Models}
\begin{document}

\onecolumn
  \icmltitle{Optimistic World Models: \\Efficient Exploration in Model-Based Deep Reinforcement Learning}



  \icmlsetsymbol{equal}{*}

  \begin{icmlauthorlist}
    \icmlcorrespondingauthor{Akshay Mete}{akshaymete@tamu.edu}
    \icmlauthor{Akshay Mete}{TAMU}
    \icmlauthor{Shahid Aamir Sheikh}{TAMU}
    \icmlauthor{Tzu-Hsiang Lin}{TAMU}
    \icmlauthor{Dileep Kalathil}{TAMU}
    \icmlauthor{P. R. Kumar}{TAMU}

  \end{icmlauthorlist}

  \icmlaffiliation{TAMU}{Department of Electrical and Computer Engineering, Texas A\&M University, College Station, Texas, USA}

  \icmlkeywords{}

  \vskip 0.3in



\printAffiliationsAndNotice{}  % no special notice (required even if empty)

\begin{abstract}
Efficient exploration remains a central challenge in reinforcement learning (RL), particularly in sparse-reward environments. We introduce Optimistic World Models (OWMs), a principled and scalable framework for optimistic exploration that brings classical reward-biased maximum likelihood estimation (RBMLE) from adaptive control into deep RL. In contrast to upper confidence bound (UCB)–style exploration methods, OWMs incorporate optimism directly into model learning by augmentation with an optimistic dynamics loss that biases imagined transitions toward higher-reward outcomes. This fully gradient-based loss requires neither uncertainty estimates nor constrained optimization. Our approach is plug-and-play with existing world model frameworks, preserving scalability while requiring only minimal modifications to standard training procedures. We instantiate OWMs within two state-of-the-art world model architectures, leading to Optimistic DreamerV3 and Optimistic STORM, which demonstrate significant improvements in sample efficiency and cumulative return compared to their baseline counterparts.
\end{abstract}
\section{Introduction}
World models such as DreamerV3 \cite{hafner2023mastering}, STORM \cite{zhang2023storm}, and DIAMOND \cite{alonso2024diffusion} have achieved state-of-the-art sample efficiency in a wide range of reinforcement learning tasks. However, they are known to underperform in sparse-reward environments as exploration strategies such as policy (actor) entropy regularization are often insufficient. 

We address this shortcoming by proposing an \textbf{Optimistic World Model} framework that introduces an \emph{optimistic dynamics loss} function on top of the standard world model training objective. This loss function \emph{gently} steers the model's transition probabilities towards higher rewards, training the world model to generate \emph{optimistic imaginations}.

\begin{figure*}[h!]
  \centering
  \begin{subfigure}[b]{0.2\linewidth}
      \includegraphics[width=\linewidth]{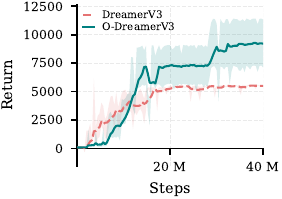}
      \caption{Private Eye }
      \label{fig:1a}
  \end{subfigure}\hfill
  \begin{subfigure}[b]{0.2\linewidth}
\includegraphics[width=\linewidth]{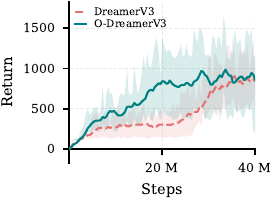}
      \caption{Enduro}
      \label{fig:1b}
  \end{subfigure}\hfill
  \begin{subfigure}[b]{0.2\linewidth}
      \includegraphics[width=\linewidth]{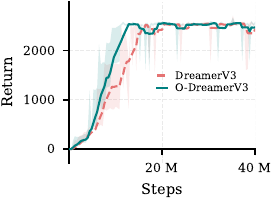}
      \caption{Montezuma's Revenge}
      \label{fig:1c}
  \end{subfigure}\hfill
  \begin{subfigure}[b]{0.2\linewidth}
\includegraphics[width=\linewidth]{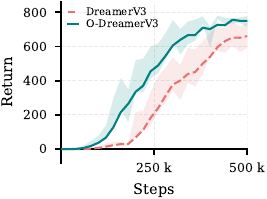}
      \caption{Cartpole Swing. Sparse}
      \label{fig:1d}
  \end{subfigure}
  \caption{Optimistic World Models on challenging environments.  }
  \label{fig:highlights}
\end{figure*}
Our Optimistic World Models (OWMs) incorporate optimistic exploration while retaining similar scalability and computational efficiency of standard world models, as the underlying neural network architectures are identical. The introduction of the optimistic dynamics loss, inspired by the Reward Biased Maximum Likelihood Estimation (RBMLE) principle \cite{kumar1982new}, avoids the computational pitfalls commonly experienced by UCB-based approaches to optimism \cite{lai1985asymptotically,auer2008near,abbasi2011regret}, namely non-convex constraints and uncertainty estimates. 

Empirically, the proposed OWM variants -- Optimistic DreamerV3 (O-DreamerV3) and Optimistic STORM (O-STORM) show improvement over DreamerV3 and STORM across many popular RL benchmarks, and achieve exceptional improvement in sparse-reward environments such as Private Eye, Acrobot Swingup Sparse, Cartpole Swingup Sparse, as well as dense-reward environments such as Enduro and UpNDown, as highlighted in Figures \ref{fig:highlights} and \ref{fig:sparse_env}.

Through this work, we also highlight that the current world model framework of separating model learning and policy learning is analogous to the certainty equivalence principle from adaptive control theory. The certainty equivalence principle is known to suffer from the closed-loop identification problem, which can lead to convergence to sub-optimal policies. This closed-loop identification problem is the underlying problem making exploration necessary in RL. RBMLE, which was notably one of the first approaches that was shown to overcome the closed-loop identification problem, forms the theoretical foundation of Optimistic World Models. 
\subsection*{Contributions}
Our key technical contributions are summarized as follows:
\begin{itemize}
\item \textbf{Gradient-Based RBMLE for Deep MBRL}: We present a fully gradient-based computational approach to implement the RBMLE principle in MBRL. This enables the application of the RBMLE principle to large-scale DRL, which has been primarily limited to small-scale RL scenarios until now.
    \item \textbf{Optimistic World Models}: We introduce Optimistic World Models, a plug-and-play framework with an additional loss function that steers the model estimate towards models with higher rewards. The OWMs incorporate principled optimistic exploration with minimal computational overload and no architectural changes. We develop two specific instances of OWMs -- Optimistic DreamerV3 and Optimistic STORM.
    \item  \textbf{Strong Empirical Gains}: OWMs exhibit exceptional improvement on challenging sparse-reward environments in the Atari100K and DMC benchmarks, highlighting their efficient exploration. Optimistic DreamerV3 achieves a mean human-normalized score (HNS) of 152.68\% -- a 55\% improvement compared to the 97.45\% mean HNS of DreamerV3. 
\end{itemize}
\section{Related Work}
\textit{Model-Based Reinforcement Learning (MBRL):} MBRL algorithms have been shown 
to achieve sub-linear regret guarantees in a variety of RL setups such as MDPs \citep{auer2008near,bourel2020tightening,mete2021reward,jin2020provably}, LQG systems \cite{abbasi2011regret,simchowitz2020naive,jedra2022minimal}, and non-linear systems \cite{kakade2020information,wagenmaker2023optimal}. Despite strong theoretical guarantees, most of these algorithms are computationally intractable or expensive. MBPO \cite{janner2019trust} and PETS \cite{chua2018deep} demonstrated how model-based algorithms can outperform model-free algorithms in deep reinforcement learning by utilizing model-generated trajectories.

\textit{World Models in Reinforcement Learning.} World models \citep{ha2018world} have gained popularity in various areas such as reinforcement learning, generative modeling, and robotics \citep{hafner2023mastering,hafner2025training,bruce2024genie,assran2025v}. In model-based reinforcement learning, world models have achieved improved sample complexity due to their ability to learn efficient representations of the environment. These world model-based RL algorithms can be broadly categorized into two categories: (a) algorithms that use policy gradient in \emph{imagination} such as the Dreamer family
\citep{hafner2020dreamerv1,hafner2022dreamerv2,hafner2023mastering}, STORM \citep{zhang2023storm}, IRIS \cite{micheli2023transformers}, and DIAMOND \citep{alonso2024diffusion}; and (b) algorithms that use Monte-Carlo Tree Search (MCTS) methods such as MuZero \citep{schrittwieser2020mastering}, EfficientZero \citep{ye2021mastering}, and EfficientZero-V2 \citep{wang2024efficientzero}. In many scenarios, the MCTS-based algorithms outperform the imagination-based algorithms at a higher computational cost, one of the underlying reasons being poor exploration by imagination-based algorithms. 

\textit{Exploration in Model Based RL.}  
World models such as Dreamer, STORM, and Diamond use exploration strategies such as policy entropy regularization, which are often found insufficient in sparse-reward environments.  The optimism in the face of uncertainty (OFU) principle, embodied by the UCB algorithms, is a popular exploration method in small-scale RL setups \cite{lai1985asymptotically,auer2002finite,auer2006logarithmic,auer2008near,abbasi2011regret}. These algorithms are computationally prohibitive in large-scale deep RL due to non-convex constraints and the requirement of precise uncertainty estimates. Several algorithms, such as H-UCRL \cite{curi2020efficient} and SOMBRL \cite{sukhija2025sombrl}, implement UCB-like optimistic exploration algorithms by utilizing Gaussian Processes to quantify model uncertainty. 
Other popular heuristic-based exploration approaches include probabilistic ensemble-based approaches \cite{chua2018deep,kurutach2018model,shyam2019model}, and the intrinsic motivation approaches \cite{burda2018exploration,pathak2017curiosity}.

\section{Preliminaries}\label{sec:background}

A Markov Decision Process (MDP) is defined by the tuple $(\gS,\gA, p^\star,r, \mu)$, where $\gS$ is the state space, $\gA$ is the action space, $p^\star: \gS \times \gA \to \Delta(\gS) $ denotes the unknown transition probability function, $r: \gS \times \gA \to \mathbb{R}$ is the reward function, and $\mu$ is the initial state distribution. Given a model $p \in P$ and  a policy $\pi \in \Pi$, the expected cumulative reward is defined as 
\begin{align}
    \label{eq:J-p-pi}
    J(p,\pi) = \E_{p,\pi} \left[ \sum_{t = 0}^{T-1} r(s_t,a_t)\mid s_0 \sim \mu \right].
\end{align}

The goal of a reinforcement learning algorithm is to find the optimal policy $\pi^\star$ such that $\pi^\star \in \arg \max J(p^\star,\pi)$ when the true model $p^\star$ is unknown. In particular, the MBRL algorithm focuses on learning the model $p^\star$ or an approximation of it using the data, and uses this learned model to learn the optimal policy. 

Recently, MBRL algorithms using the framework of the world model, such as Dreamer  \citep{hafner2023mastering,hafner2022dreamerv2,hafner2020dreamerv1}, STORM \citep{zhang2023storm}, DIAMOND \citep{alonso2024diffusion}, and IRIS \citep{micheli2023transformers}, have shown remarkable successes in many challenging tasks. The model and policy learning in this class of algorithms follows the following general framework. 
\begin{enumerate}
    \item \textbf{Environment Interaction:} The agent deploys its current policy in the real environment, and stores observations ($o_t$), actions ($a_t$), and rewards ($r_t$) in a replay buffer.
    \item \textbf{Training the world model:} The world model is trained to learn an accurate representation of the environment by fitting on data stored in the replay buffer. Typically, the world model includes an encoder that maps observations ($o_t$) to a latent space representation ($s_t$), a dynamics model that predicts the next state given the current state and action ($p(s_{t+1}|s_t,a_t)$), a reward model ($r(s_t,a_t)$), and a decoder to reconstruct the observation from the latent space representation.
    \item \textbf{Learning in imagination:} The current world model and policy ($\pi(a_t|s_t)$) are used to generate \emph{imagined} trajectories. These imagined trajectories are used to improve the policy using methods such as actor-critic algorithms.
\end{enumerate}

These steps are performed iteratively to improve both the world model of the environment and the agent's policy. In this framework, the world model is trained on the real data, while the actor is only trained on the imaginary trajectories.  This separation between world model and policy learning is analogous to the certainty equivalence principle described below.

\subsection{Bias of MLE}
For fully observable MDPs ($o_t=s_t$), the certainty equivalence approach to MBRL/adaptive control is as follows: 
The learning agent computes the maximum likelihood estimate,
\begin{align*}
    p_{t} \in \arg \max_{p \in P} \sum_{\tau=0}^{t-1} \log p(s_{\tau+1} \mid s_\tau,a_\tau).
\end{align*}
Then it selects a policy  $\pi_t \in \Pi$ that is optimal for the estimated model $p_t$:
\begin{equation*}
    \pi_t \in \arg \max_{\pi \in \Pi} J(p_t,\pi) .
\end{equation*}
A natural question is whether the certainty equivalence approach always converges to the true dynamics $p^\star$ and the optimal policy $\pi^\star$? 
This fundamental question was first studied in \cite{mandl1974estimation}, and it was shown that $p_t$ converges to $p^\star$ if the following condition holds:
\begin{equation*}
   \text{ For any } p^1 \neq p^2 \in P,  p^1(\cdot \mid s,a) \neq p^2(\cdot \mid s,a) \text{ for all } s,a.
\end{equation*}
This requirement, known as the ``identifiability condition'', is highly restrictive, and is not satisfied even in simple settings such as multi-armed bandits.
\citet{borkar1979adaptive} further showed that in the absence of this identifiability condition, the parameter identification only occurs for the closed-loop system, i.e.,
\begin{equation}
p_\infty(s'\mid s,\pi_\infty(s))=p^\star(s'\mid s,\pi_\infty(s)) ~\forall ~ s,s'
\end{equation}
where $p_\infty$ and $\pi_\infty$ are the limiting estimate and policy, respectively. This is known as the ``closed-loop identification''. 

However, this does not guarantee that $p_\infty(s'\mid s,a)=p^\star(s'\mid s,a) ~\forall ~ s,a,s' 
$, nor does it guarantee that $\pi_\infty$ is optimal for $p^\star$.
This central issue is known as the ``closed-loop identification problem'' in adaptive control. 

As a consequence, $p_t$ and $\pi_t$ can converge to a model $p_\infty$ and its optimal policy $\pi_\infty$ that are self-consistent but suboptimal for the true model \cite{borkar1979adaptive}: 
\begin{equation*}
    J(p_\infty,\pi_\infty) = J(p^\star,\pi_\infty) \implies J(p^\star,\pi_\infty) \le J(p^\star,\pi^\star).
\end{equation*}
This bias of the maximum likelihood estimate toward models with lower optimal rewards underscores the need for exploration, making the exploration-exploitation trade-off a fundamental challenge in reinforcement learning. 

\subsection{Reward Biased Maximum Likelihood Estimation}
One of the first solutions to the above closed-loop identifiability issue, the \textit{reward-biased} maximum likelihood estimation (RBMLE) method, was proposed by \citet{kumar1982new}. To overcome the inherent bias of MLE, RBMLE introduces an explicit, counter-acting bias towards models with higher rewards as follows:
\begin{equation}
 p_t \in \arg \max_{p \in P}~ \left\{\alpha(t) \max_{\pi \in \Pi}J(p,\pi) + \frac{1}{t} \sum_{\tau=0}^{t-1} \log p(s_{\tau+1} \mid s_\tau,a_\tau)\right\}.\label{eq:rbmle_orig}
\end{equation}
\begin{equation*}
    \pi_t \in \arg \max_{\pi \in \Pi} J(p_t,\pi),
\end{equation*}
The term $\alpha(t)$ has to be carefully designed. It should be small enough to preserve the good properties of the maximum likelihood estimate, i.e., closed-loop identification. At the same time, it has to be large enough to overcome the bias of MLE towards models with lower average rewards. Both of these properties are achieved by choosing $\alpha(t)$ such that $\lim_{t \to \infty }t\alpha(t) \to \infty $ and $\lim_{t \to \infty }\alpha(t) \to 0 $. \citet{kumar1982new} showed that the model estimate $p_t$ converges to $p^\star$ and the policy $\pi_t$ to an optimal policy, in a Cesaro sense.

The RBMLE principle has been adapted and studied in a wide range of adaptive control/MBRL scenarios, such as stochastic bandits, LQG control, and RLHF \cite{becker1981optimal,kumar1982new, campi1996optimal,duncan1994almost,liu2023maximize,hung2021reward,mete2021reward,mete2022augmented,hung2023value,cen2024value}. A detailed account of various works based on the RBMLE principle can be found in \cite{mete2022rbmle,mete2023reward}.

In the next section, we provide a computationally efficient method to implement the RBMLE principle, which has been a challenge so far. We then propose the Optimistic World Models based on it.
\begin{figure*}[htb!]
  \centering
\includegraphics[width=0.9\linewidth]{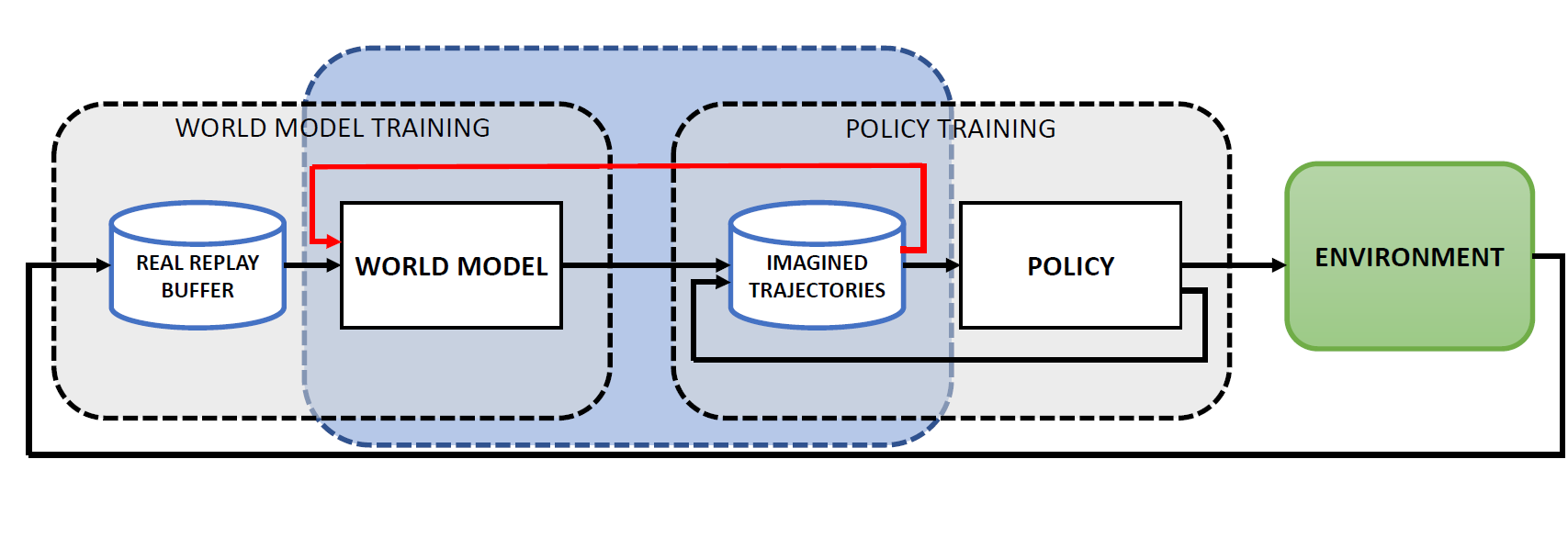}
  \caption{ Optimistic World Model framework: In standard world models, the dynamics model is trained only to fit the real replay buffer. In OWMs, the optimistic dynamics loss uses imagined trajectories to push model transitions toward high-reward outcomes (highlighted by the red arrow), leading to \emph{optimistic imaginations}.
  }
\label{fig:opt_world_model_framework}\end{figure*}

\section{Optimistic World Models} \label{sec:owm}
In this section, we introduce \emph{Optimistic world models}, a principled optimistic exploration framework for world models.
We begin by introducing a gradient based computational approach to adapt the RBMLE approach in deep reinforcement learning.

\subsection{RBMLE in Deep Reinforcement Learning} \label{sec:rbmle_drl}

Consider a model-based reinforcement learning agent with two neural networks: A \emph{model} network, parameterized by $\phi$, that learns $p_\phi(s'\mid s,a)$, the transition probabilities of the next state $s'$ given current state $s$ and action $a$, and a \emph{policy} network parameterized by $\theta$ that defines the policy $\pi_\theta(a|s)$. Let $D_t=\{(s_u,a_u,s_{u+1})\}_{u=1}^{t-1}$ be the replay buffer, which stores tuples collected from real environment interaction. The RBMLE objective is given by:
\begin{equation}
M_{\phi, \theta}=\alpha(t)J(p_\phi,\pi_\theta)+ \frac{1}{t} \sum_{(s,a,s') \in D_t}{\log p_\phi(s' \mid s,a)}.
\end{equation}
Let $\tau_{\phi,\theta} = \{s_0,a_0,\cdots,s_{H+1}\}$ denote an imagined trajectory generated by simulating the policy $\pi_\theta$ on the model $p_\phi$, 
then the gradient of $J(p_\phi,\pi_\theta)$ can be computed using the log-derivative trick similar to the vanilla policy gradient \cite{sutton1999policy}, and is given by: 
\begin{equation}
    \nabla_\phi J(p_\phi,\pi_\theta) =\E_{\tau_{\phi,\theta}}\left[\sum_{h=0}^H R(\tau_{\phi,\theta}) \nabla_\phi \log p_\phi(s_{h+1} \mid s_h,a_h)  \right].
\end{equation}
%============================

In practice, this can be empirically computed by generating $N$ trajectories $\{\tau^i_{\phi,\theta}\}_{i=1}^N$, and then computing the sample mean.
\begin{align}
    &\nabla_\phi J(p_\phi,\pi_\theta)  =\frac{1}{N}\sum_{i=1}^N\left[\sum_{h=0}^H R(\tau^i_{\phi,\theta}) \nabla_\phi \log p_\phi(s_{h+1} \mid s_h,a_h)  \right].    
\end{align}
This yields us the gradient estimate of $M_{\phi,\theta}$,
\begin{align}\label{eq:model_grad}
\nabla_\phi M_{\phi,\theta} = &\frac{\alpha(t)}{N}\left[\sum_{i=1}^N \sum_{h=0}^H R(\tau^i_{\phi,\theta}) \nabla_\phi \log p_\phi(s_{h+1} \mid s_h,a_h)  \right] + \frac{1}{t}\sum_{(s,a,s') \in D_t}{\nabla_\phi \log p_\phi(s' \mid s,a)}.
\end{align}  
Similarly,  an estimate of $\nabla_\theta J(p_\phi,\pi_\theta)$ can be computed using the imagined trajectories.
\begin{equation} \label{eq:policy_grad}
    \nabla_\theta J(p_\phi,\pi_\theta) =\frac{1}{N}\sum_{i=1}^N\left[\sum_{h=0}^H R(\tau^i_{\phi,\theta}) \nabla_\theta \log \pi_\theta(a_h \mid s_h)  \right].
\end{equation}
Therefore, (\ref{eq:model_grad}) and
(\ref{eq:policy_grad}) can be used to update the model and policy parameters as follows:
\begin{align*}
    \phi_{t+1} = \phi_t + \beta  \nabla_\phi  M_{\phi,\theta}|_{\phi_t,\theta_t},~ 
    \theta_{t+1} =\theta_t + \beta  \nabla_\theta  J(p_\phi,\pi_\theta)|_{\phi_t,\theta_t}.
\end{align*}
The key insights from the RBMLE approach are:
\begin{itemize}
    \item \textbf{Optimistic Dynamics Loss:} RBMLE can be implemented by augmenting the certainty equivalence framework with an additional loss term. This loss function steers the log-probabilities of the model transitions, $\log p_\phi(s'|s,a)$ towards higher reward outcomes, analogous to how policy gradient methods push $\log \pi_\theta(a \mid s)$ towards actions with higher rewards. 
    \item \textbf{The optimism function $\alpha(t)$:} The parameter $\alpha(t)$ controls the degree of optimism by balancing the optimistic dynamics loss and the log-likelihood loss. 
\end{itemize}
 Based on these insights, we develop Optimistic World Models, generalizing the RBMLE principle to deep reinforcement learning for partially observable MDPs.   \subsection{Incorporating Optimism in World Models}
We introduce the optimistic dynamics loss, $\gL_t^{opt}$ in the imagination-based world model framework as follows: Let $\tau_t = \{(s_\ell,a_\ell,r_\ell)\}_{\ell=1}^{L}$ denote the \emph{imagined} trajectory generated by executing the current policy $\pi_{\theta_t}$ on the learned world model dynamics $p_{\phi_t}$. 
\begin{align}
 \label{eq:l_opt}
    \gL_t^{opt}= -&\alpha(t)\left[\sum_{\ell=1}^{L-1} A_\ell \log p_{\phi_t}(s_{\ell+1} \mid s_\ell,a_\ell) \right] -\eta \sum_{\ell=0}^{L-1} H(p_{\phi_t}(s_{\ell+1} \mid s_\ell,a_\ell)) ~~,~~   
\end{align}
where $A_\ell$ is the advantage function computed using $r_\ell$, the value $V(s_\ell)$ given by the critic network, and $H$ is the entropy function of the distribution $p_{\phi_t}(\cdot \mid s_\ell,a_\ell)$. 

This optimistic dynamics loss improves upon (\ref{eq:model_grad}) by utilizing the advantage function $A_\ell$ and the additional model entropy term $H$. These improvements mirror the advances in policy gradient algorithms to address issues such as high variance and entropy collapse. As a consequence, the optimistic dynamics loss is structurally the same as that of the actor loss and is simply obtained by substituting $\log \pi_\theta(a_\ell|s_\ell)$  with $\log p_\phi(s_{\ell+1}|s_\ell,a
_\ell)$. 

The OWM framework can be adapted to any world model based on the framework described in Section \ref{sec:background}, such as Dreamer \citep{hafner2023mastering}, STORM \citep{zhang2023storm}, TWM \citep{robine2023transformer}, IRIS \cite{micheli2023transformers}, and DIAMOND \cite{alonso2024diffusion}.
In this paper, we propose two instances of OWMs: Optimistic DreamerV3 and Optimistic STORM.
\subsubsection{Optimistic DreamerV3}
We introduce Optimistic DreamerV3, the optimistic variant of DreamerV3. DreamerV3 is one of the leading MBRL algorithms known for its high sample efficiency without requiring task-specific hyperparameter tuning. 
Beyond the addition of the optimistic dynamics loss, Optimistic DreamerV3 is identical to the original DreamerV3; all other components and learning objectives remain unchanged. 

The imagined state $s_\ell$ is comprised of a stochastic component $z_\ell$ and a deterministic component $h_\ell$. Therefore, we replace $p_\phi(s_{\ell+1} \mid s_\ell,a_\ell)$ with $p_\phi(z_{\ell+1} \mid h_{\ell+1})$. The optimistic dynamics loss term in Optimistic DreamerV3 is: 
\begin{align}
    \gL_t^{opt} = &-\alpha(t)\left[\sum_{\ell=0}^{L-1} A_\ell\log p_\phi(z_{\ell+1} \mid h_{\ell+1}) \right]- \eta  \sum_{\ell=0}^{L-1} H \left(p_\phi(z_{\ell+1} \mid h_{\ell+1})\right).    
\end{align}
The advantage term $A_\ell$, is computed as $\left(\frac{G_\ell^\lambda - V(s_\ell)}{\max\{1,S\}}\right)$. Here, $G_\ell^\lambda$ is the $\lambda$-return, $V$ is the value function estimated by the critic, $\eta$ is an entropy coefficient, $H$ is the entropy function, and $S$ is an exponential moving average of the range between the 5th and 95th percentiles of returns, defined as,
$S=EMA(Per(G_\ell^\lambda,95)-Per(G_\ell^\lambda,5),0.99).
$

\subsubsection{Optimistic STORM}
We propose Optimistic STORM (O-STORM), the optimistic variant of the STORM world model. \citet{zhang2023storm} shows that the transformer-based STORM world model outperforms the DreamerV3 algorithm in many Atari games in the Atari100K benchmark. 

The optimistic dynamics loss function is computed as: 
\begin{align}
    \gL_t^{opt} = -&\alpha(t)\left[\sum_{\ell=0}^{L-1}A_\ell\log p_\phi(s_{\ell+1} \mid s_\ell,a_\ell)\right]\nonumber-\eta \sum_{\ell=0}^{L-1} H(p_\phi(s_{\ell+1} \mid s_\ell,a_\ell)). 
\end{align}
The definition of the advantage function, $A_\ell$, is identical to the one from Optimistic DreamerV3. All other components and learning objectives of Optimistic STORM are identical to those of STORM and can be found in \cite{zhang2023storm}. 

%===========================================
\begin{figure*}[t!]
  \centering
  % First row --------------------------------------------------------------
  \begin{subfigure}[b]{0.22\linewidth}
      \includegraphics[width=\linewidth]{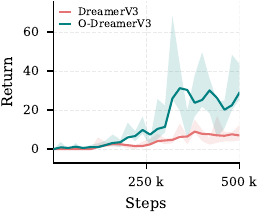}
      \caption{Acrobot Swingup Sparse}
      \label{fig:3a}
  \end{subfigure}\hfill
  \begin{subfigure}[b]{0.22\linewidth}
      \includegraphics[width=\linewidth]{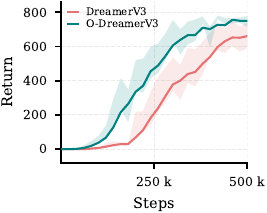}
      \caption{Cartpole Swingup Sparse}
      \label{fig:3b}
  \end{subfigure}\hfill
  \begin{subfigure}[b]{0.22\linewidth}
      \includegraphics[width=\linewidth]{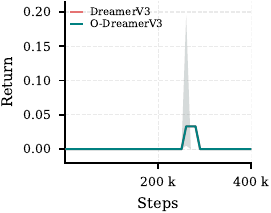}
      \caption{Freeway }
      \label{fig:3c}
  \end{subfigure}\hfill
  \begin{subfigure}[b]{0.22\linewidth}
      \includegraphics[width=\linewidth]{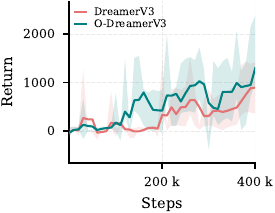}
      \caption{Private Eye}
      \label{fig:3d}
  \end{subfigure}
  
  \vspace{4pt} % small vertical gap between the two rows
  
  % Second row -------------------------------------------------------------
  \begin{subfigure}[b]{0.22\linewidth}
      \includegraphics[width=\linewidth]{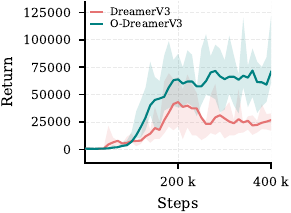}
      \caption{UpNDown}
      \label{fig:3e}
  \end{subfigure}\hfill
  \begin{subfigure}[b]{0.22\linewidth}
      \includegraphics[width=\linewidth]{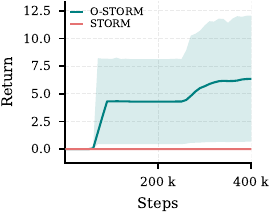}
      \caption{Freeway (O-STORM)}
      \label{fig:3f}
  \end{subfigure}\hfill
  \begin{subfigure}[b]{0.22\linewidth}
      \includegraphics[width=\linewidth]{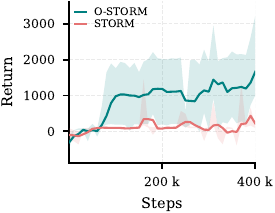}
      \caption{ Private Eye (O-STORM) }
      \label{fig:3g}
  \end{subfigure}\hfill
  \begin{subfigure}[b]{0.22\linewidth}
      \includegraphics[width=\linewidth]{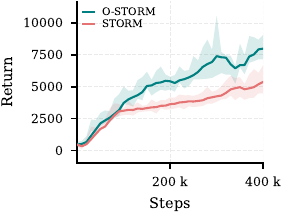}
      \caption{ UpNDown (O-STORM)}
      \label{fig:3h}
  \end{subfigure}
  \caption{Optimistic World Models on sparse reward environments. }
  \label{fig:sparse_env}
\end{figure*}
%===========================================
\section{Experiments} 
We evaluate the OWM variants: Optimistic DreamerV3 and Optimistic STORM on the widely adopted Atari100K and DeepMind Control (DMC) benchmarks. Our main results are presented in this section; comprehensive learning curves, interquartile mean (IQM) and median scores, ablations, and experimental details are provided in Appendix \ref{app:exp}. All learning curves are reported as mean reward $\pm$ the standard error of the mean. The results for O-DreamerV3 on the Atari100K and the DMC benchmarks are averaged over 10 seeds. All other results, including those for Optimistic STORM and the ablation studies, are averaged over 5 seeds due to computational resource limitations.

\subsection*{Performance on Sparse Reward Environments}
Figure \ref{fig:sparse_env} highlights the performance of OWMs on challenging, sparse-reward environments, including Private Eye, Freeway, Acrobot Swingup Sparse, and Cartpole Swingup Sparse. OWMs outperform baselines on these sparse environments, where standard world models suffer due to a lack of efficient exploration. They also outperform baselines on dense-reward environments such as UpNDown.

Additionally, Figure \ref{fig:highlights} shows the long-term performance (40 million samples) of O-DreamerV3 and DreamerV3 on hard exploration Atari games, Private Eye and Montezuma's Revenge. The O-DreamerV3 achieves nearly 2X mean return compared to DreamerV3 on Private Eye at 40M samples. While it achieves the same score as DreamerV3 on Montezuma's Revenge with fewer samples. 

\subsection*{Performance on Atari100K Benchmark}
The Atari100K benchmark is designed to test the performance of RL algorithms in a limited sample regime. Optimistic DreamerV3 achieves a mean human-normalized score (HNS) of $152.68\%$ on the Atari100K benchmark, compared to $97.45\%$ mean HNS of DreamerV3.

O-STORM outperforms STORM in sparse-reward games such as Private Eye and Freeway, notably achieving a positive score on Freeway, unlike STORM, DreamerV3, and O-DreamerV3. O-STORM achieves a mean HNS of $80.68\%$, while STORM achieves a mean HNS of $75.90\%$.

\subsection*{Performance on DMC Suite}
\begin{figure*}[h!] 
    \centering
    % --- First Subfigure (Top) ---
    \begin{subfigure}[b]{0.74\textwidth}
        \centering
        \includegraphics[width=\linewidth]{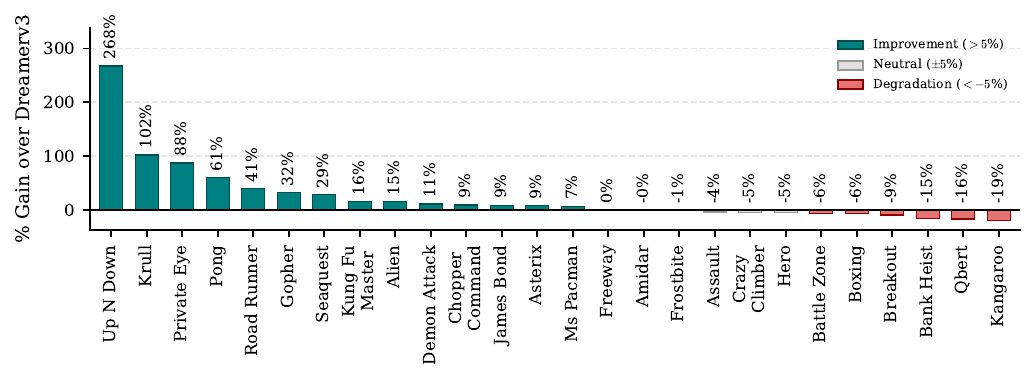}
        \caption{Optimistic DreamerV3 on Atari100K}
        \label{fig:odreamer_atari_bar}
    \end{subfigure}
             \begin{subfigure}[b]{0.74\textwidth}
        \centering
        \includegraphics[width=\linewidth]{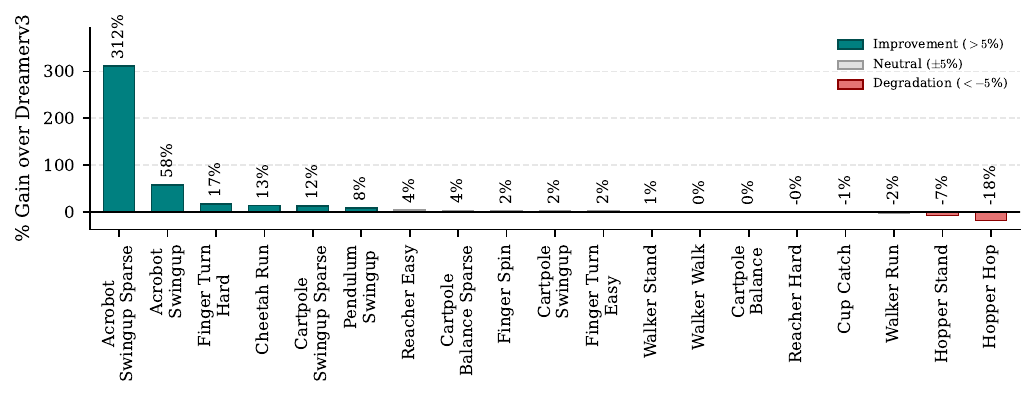}
        \caption{Optimistic DreamerV3 on DMC Proprio}
        \label{fig:odreamer_dmc_proprio_bar}
    \end{subfigure}    
    \begin{subfigure}[b]{0.74\textwidth}
        \centering
        \includegraphics[width=\linewidth]{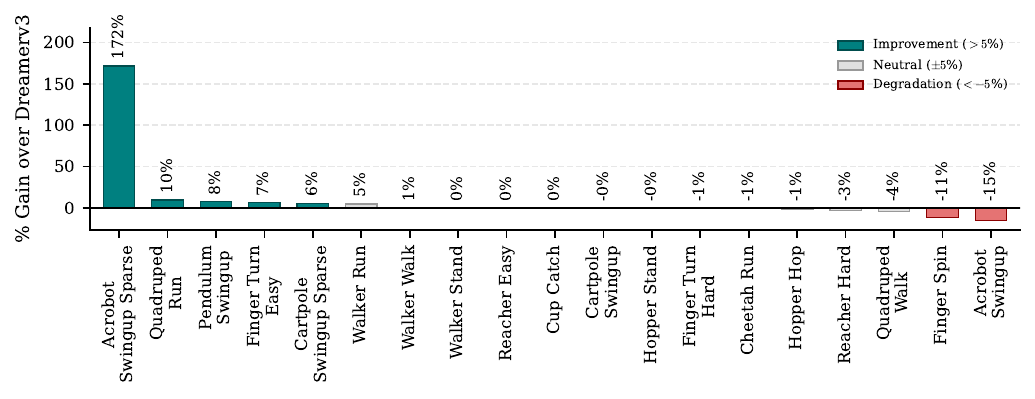}
        \caption{Optimistic DreamerV3 on DMC Vision}
        \label{fig:odreamer_dmc_vision_bar}
    \end{subfigure}   
    \begin{subfigure}[b]{0.74\textwidth}
        \centering
        \includegraphics[width=\linewidth]{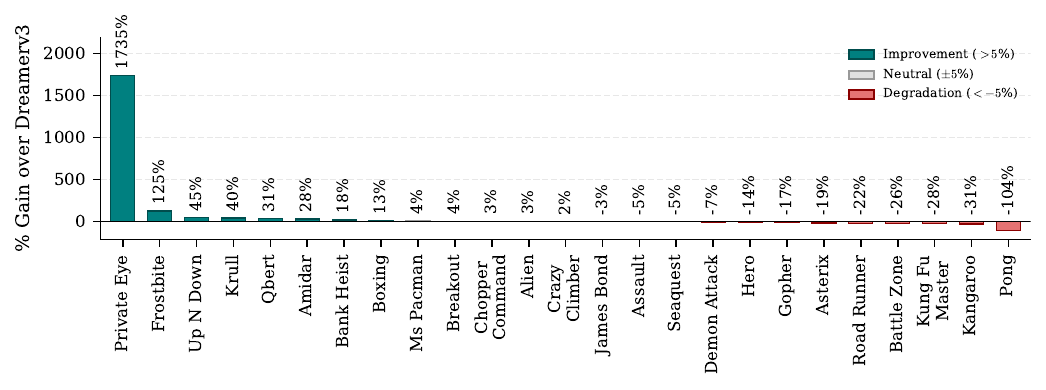}
        \caption{Optimistic STORM on Atari100K}
        \label{fig:ostorm_atari_bar}
    \end{subfigure}
    \caption{\textbf{Performance of Optimistic World Models:} We plot the \% gain of Optimistic DreamerV3 and Optimistic STORM over DreamerV3 and STORM respectively. Freeway is not included in Figure \ref{fig:ostorm_atari_bar} as the baseline STORM has a score of $0$, while O-STORM achieves a mean score of $6.38$. }
    \label{fig:atari_bar}
\end{figure*}

We further evaluate the efficiency of Optimistic DreamerV3 on the DeepMind Control (DMC) suite. The DMC suite contains the DMC Proprio benchmark, which uses state-based inputs, and the DMC Vision benchmark, which uses image-based inputs. While many of the tasks in DMC benchmarks are easy, and both algorithms have similar performance on most DMC environments, O-DreamerV3 improves the performance on sparse-reward environments such as Cartpole Swingup Sparse, and Acrobot Swingup Sparse. Figures \ref{fig:odreamer_dmc_proprio_bar} and \ref{fig:odreamer_dmc_vision_bar} summarize the performance on DMC Proprio and DMC Vision, respectively.
\clearpage
\clearpage
\subsection*{The optimistic bias term $\alpha(t)$}
\begin{figure*}[t!]
  \centering
  % First row --------------------------------------------------------------
  \begin{subfigure}[b]{0.22\linewidth}
      \includegraphics[width=\linewidth]{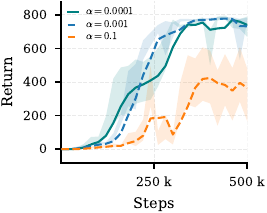}
      \caption{$\alpha$}
      \label{fig:5a}
  \end{subfigure}\hfill
  \begin{subfigure}[b]{0.22\linewidth}
      \includegraphics[width=\linewidth]{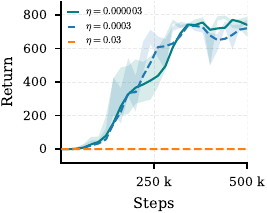}
      \caption{$\eta$}
      \label{fig:5b}
  \end{subfigure}\hfill
  \begin{subfigure}[b]{0.22\linewidth}
      \includegraphics[width=\linewidth]{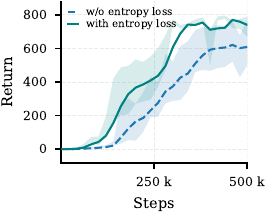}
      \caption{Entropy Loss}
      \label{fig:5c}
  \end{subfigure}
    \begin{subfigure}[b]{0.22\linewidth}
      \includegraphics[width=\linewidth]{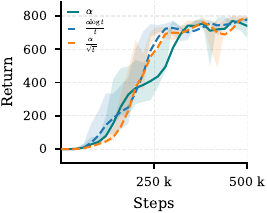}
      \caption{$\alpha(t)$}
      \label{fig:5d}
  \end{subfigure}\hfill
\caption{Ablations and hyperparameter sensitivity results for Cartpole Swingup Sparse (DMC Proprio): (a) Sensitivity to the optimism term $\alpha$: poor performance at $\alpha=0.1$ shows that optimism should be \emph{mild}. (b) Sensitivity to the model entropy loss coefficient $\eta$: no learning at $\eta=0.03$, while smaller values are beneficial. (c) Performance without the entropy loss: the entropy loss improves the performance of O-DreamerV3. (d) Performance under various decay schedules of $\alpha(t)$. Ablations for additional environments are provided in the Appendix.}
  \label{fig:main_ablations}
\end{figure*}

OWMs introduce two additional parameters, namely, the optimistic bias term $\alpha(t)$ and the model entropy loss coefficient $\eta$. Similar to DreamerV3, O-DreamerV3 also enjoys a single hyperparameter configuration across all tasks and benchmarks. We used $\alpha(t)=\alpha = 0.0001$ as a constant hyperparameter.  Figure \ref{fig:5d} shows the performance for different decay schedules of $\alpha(t)$. 
The entropy loss coefficient $\eta$ is set to be $3 \times 10^{-6}$ for O-DreamerV3 and $3\times10^{-4}$ for O-STORM. Ablations on the entropy loss term in Figure \ref{fig:5b} and \ref{fig:5c} highlight the gain due to the entropy loss term.

The ablations and sensitivity analysis shown in Figures \ref{fig:main_ablations}, \ref{fig:eta_ablations}, \ref{fig:alpha_sensi} and \ref{fig:eta_sensi} provide a clear guideline on choosing $\alpha$ and $\eta$. They suggest that high values such as  $\alpha=0.1$ or $\eta=0.03$ can degrade performance drastically, while smaller values are beneficial.

\subsection*{Computational Overload}
OWMs are plug-and-play modifications over standard world models, adding marginal computational overhead compared to baselines, as shown in Table \ref{tab:compute_time}.
\begin{table}[h]
  \caption{Training time (minutes) for various algorithms on Atari100K (MsPacman) using an RTX 4090 GPU, averaged over 5 seeds.}
  \label{tab:compute_time}
  \begin{center}
    \begin{small}
      \begin{sc}
        \begin{tabular}{cccc}
          \toprule
           O-DreamerV3         & DreamerV3      & O-STORM  & STORM\\
          \midrule 138
 & 115   & 178  &   170  \\

          \bottomrule
        \end{tabular}
      \end{sc}
    \end{small}
  \end{center}
\end{table}
\section{Discussion}\label{sec:discussion}
RBMLE, UCB, and Thompson Sampling are three foundational principles for exploration studied in MBRL/adaptive control. RBMLE, first proposed in \cite{kumar1982new}, and UCB, first proposed in \cite{lai1985asymptotically}, embody the principle of optimism as described below.
\subsection*{A Tale of Two Approaches to Optimism: RBMLE and UCB} 
For the MDP setup described in Section \ref{sec:background}, the general UCB principle chooses the model estimate $p_t$ with the highest potential average reward within a confidence interval, as follows:
\begin{align}
    p_t \in &\arg \max_p J^\star(p) \label{eq:ucb}\\
    &\text{subject to}: D(p,\hat{p}_t) \le \gC_t(\delta), \nonumber
\end{align}
where $D$ is a measure of divergence from the MLE model $\hat{p}_t$, and $\gC_t(\delta)$ is a high-probability confidence interval.

In this same scenario, RBMLE (\ref{eq:rbmle_orig}) can be equivalently seen as balancing the reward and fitting error of the model, i.e.,
\begin{align}
   p_t \in \arg \max_p \alpha(t)J^\star(p) -   D(p,\hat{p}_t) .\label{eq:rbmle}
\end{align}
UCB incorporates optimism by solving the primal optimization problem, and controls the degree of optimism by $\gC_t$. RBMLE incorporates optimism by solving its Lagrangian dual formulation, and controls the degree of optimism by $\alpha(t)$. 

RBMLE has been shown to outperform UCB and TS variants in various small-scale RL setups such as stochastic bandits \cite{liu2020exploration}, contextual bandits \cite{hung2021reward}, tabular MDPs \cite{mete2021reward}, and LQG systems \cite{mete2022augmented}. Similar approaches have been studied recently in different RL setups, such as RL with function approximation \cite{liu2023maximize}, RLHF \cite{cen2024value}, as well as model-free RL \cite{liu2023maximize}. Several works \cite{bourel2020tightening, fruit2020improved} have highlighted the excessive optimism of earlier variants of UCB algorithms, such as UCRL \cite{auer2006logarithmic} and UCRL2 \cite{auer2008near}, due to weak confidence intervals. UCRL2B \cite{fruit2020improved} and UCRL3  \cite{bourel2020tightening} utilize sharper confidence intervals to curb the excessive optimism. In contrast, RBMLE incorporates \emph{mild} optimism with small values of $\alpha(t)$.  

\subsection*{Objective Mismatch in MBRL} 
Objective mismatch is another issue of the certainty equivalence-based MBRL discussed in \cite{lambert2020objective, eysenbach2022mismatched}. 
It is attributed to the fact that the model is trained
using one objective of likelihood maximization, but the policy is trained using a different objective of average reward maximization \cite{eysenbach2022mismatched}. 
The optimistic RBMLE objective (\ref{eq:rbmle}) provides an explicit mechanism for balancing the two objectives of model accuracy and the cumulative rewards. 

\subsection*{Limitations and Future Work}

The optimistic dynamics loss provides an optimistic direction for exploration, and the step size is controlled by the optimistic bias term $\alpha(t)$. The theoretical results on RBMLE in tabular MDPs show that $t\alpha(t) \to \infty$ and $\alpha(t) \to 0$ guarantee asymptotic convergence to an optimal policy. 
Choosing $\alpha(t)=\alpha$, or even $\alpha(t)=\frac{\alpha}{\sqrt{t}}$, is perhaps too simplistic to maximize exploration gains. An adaptive, sample-path dependent design of $\alpha(t)$, similar to the meta-controller used for choosing the exploration coefficient in Agent57 \cite{badia2020agent57}, can further enhance the performance of OWMs.

The regret bounds of RBMLE have been established in different settings \cite{mete2021reward, mete2022augmented, hung2021reward, liu2020exploration, cen2024value}. In this paper, we have presented a completely gradient-based approach to RBMLE.  The convergence analysis of the practical, gradient-based RBMLE is a natural next step.
\subsection*{Conclusion}
We have introduced \textbf{Optimistic World Models}, a principled framework for efficient exploration in model-based reinforcement learning. The improved performance of OWMs compared to baselines, particularly in sparse-reward environments, validates the efficient exploration of OWMs. The simple, plug-and-play nature of OWMs allows their adaptation to many world models.

\subsection*{Acknowledgment}
This material is based upon work partially supported by the National Science Foundation under Contract Number CNS-2328395, the US Army Contracting Command under Contract Numbers W911NF2520046, W911NF2210151, and W911NF2120064, and the US Office of Naval Research under contract N00014-24-12615. This research was conducted using the advanced computing resources provided by the Texas A\&M High Performance Research Computing.

\bibliography{ref}

@article{kumar1982new,
  title={A new family of optimal adaptive controllers for {M}arkov chains},
  author={Kumar, P. R. and Becker, A},
  journal={IEEE Transactions on Automatic Control},
  volume={27},
  number={1},
  pages={137--146},
  year={1982},
  publisher={IEEE}
}

@article{borkar1979adaptive,
  title={Adaptive control of {M}arkov chains, {I}: Finite parameter set},
  author={Borkar, Vivek and Varaiya, P},
  journal={IEEE Transactions on Automatic Control},
  volume={24},
  number={6},
  pages={953--957},
  year={1979},
  publisher={IEEE}
}

@article{hafner2023mastering,
  title={Mastering diverse domains through world models},
  author={Hafner, Danijar and Pasukonis, Jurgis and Ba, Jimmy and Lillicrap, Timothy},
  journal={arXiv preprint arXiv:2301.04104},
  year={2023}
}

@article{zhang2023storm,
  title={{STORM}: Efficient stochastic transformer based world models for reinforcement learning},
  author={Zhang, Weipu and Wang, Gang and Sun, Jian and Yuan, Yetian and Huang, Gao},
  journal={Advances in Neural Information Processing Systems},
  volume={36},
  pages={27147--27166},
  year={2023}
}

@article{alonso2024diffusion,
  title={Diffusion for world modeling: Visual details matter in {A}tari},
  author={Alonso, Eloi and Jelley, Adam and Micheli, Vincent and Kanervisto, Anssi and Storkey, Amos J and Pearce, Tim and Fleuret, Fran{\c{c}}ois},
  journal={Advances in Neural Information Processing Systems},
  volume={37},
  pages={58757--58791},
  year={2024}
}

@article{ha2018world,
  title={{W}orld {M}odels},
  author={Ha, David and Schmidhuber, J{\"u}rgen},
  journal={arXiv preprint arXiv:1803.10122},
  volume={2},
  number={3},
  year={2018}
}

@article{mandl1974estimation,
  title={Estimation and control in {M}arkov chains},
  author={Mandl, Petr},
  journal={Advances in Applied Probability},
  volume={6},
  number={1},
  pages={40--60},
  year={1974},
  publisher={Cambridge University Press}
}

@misc{micheli2023transformers,
      title={Transformers are Sample-Efficient World Models}, 
      author={Vincent Micheli and Eloi Alonso and François Fleuret},
      year={2023},
      eprint={2209.00588},
      archivePrefix={arXiv},
      primaryClass={cs.LG},
      url={https://arxiv.org/abs/2209.00588}, 
}

@misc{hafner2020dreamerv1,
      title={Dream to Control: Learning Behaviors by Latent Imagination}, 
      author={Danijar Hafner and Timothy Lillicrap and Jimmy Ba and Mohammad Norouzi},
      year={2020},
      eprint={1912.01603},
      archivePrefix={arXiv},
      primaryClass={cs.LG},
      url={https://arxiv.org/abs/1912.01603}, 
}

@misc{hafner2022dreamerv2,
      title={Mastering {A}tari with Discrete World Models}, 
      author={Danijar Hafner and Timothy Lillicrap and Mohammad Norouzi and Jimmy Ba},
      year={2022},
      eprint={2010.02193},
      archivePrefix={arXiv},
      primaryClass={cs.LG},
      url={https://arxiv.org/abs/2010.02193}, 
}

@article{hafner2025training,
  title={Training agents inside of scalable world models},
  author={Hafner, Danijar and Yan, Wilson and Lillicrap, Timothy},
  journal={arXiv preprint arXiv:2509.24527},
  year={2025}
}

@article{lambert2020objective,
  title={Objective mismatch in model-based reinforcement learning},
  author={Lambert, Nathan and Amos, Brandon and Yadan, Omry and Calandra, Roberto},
  journal={arXiv preprint arXiv:2002.04523},
  year={2020}
}

@article{lai1985asymptotically,
title = {Asymptotically efficient adaptive allocation rules},
journal = {Advances in Applied Mathematics},
volume = {6},
number = {1},
pages = {4-22},
year = {1985},
issn = {0196-8858},
doi = {https://doi.org/10.1016/0196-8858(85)90002-8},
url = {https://www.sciencedirect.com/science/article/pii/0196885885900028},
author = {T.L Lai and Herbert Robbins}
}

@article{auer2002finite,
  title={Finite-time analysis of the multiarmed bandit problem},
  author={Auer, Peter and Cesa-Bianchi, Nicolo and Fischer, Paul},
  journal={Machine learning},
  volume={47},
  number={2},
  pages={235--256},
  year={2002},
  publisher={Springer}
}

@article{auer2008near,
  title={Near-optimal regret bounds for reinforcement learning},
  author={Auer, Peter and Jaksch, Thomas and Ortner, Ronald},
  journal={Advances in neural information processing systems},
  volume={21},
  year={2008}
}

@inproceedings{abbasi2011regret,
  title={Regret bounds for the adaptive control of linear quadratic systems},
  author={Abbasi-Yadkori, Yasin and Szepesv{\'a}ri, Csaba},
  booktitle={Proceedings of the 24th Annual Conference on Learning Theory},
  pages={1--26},
  year={2011},
  organization={JMLR Workshop and Conference Proceedings}
}

@inproceedings{jin2020provably,
  title={Provably efficient reinforcement learning with linear function approximation},
  author={Jin, Chi and Yang, Zhuoran and Wang, Zhaoran and Jordan, Michael I},
  booktitle={Conference on learning theory},
  pages={2137--2143},
  year={2020},
  organization={PMLR}
}

@inproceedings{mete2021reward,
  title={Reward biased maximum likelihood estimation for reinforcement learning},
  author={Mete, Akshay and Singh, Rahul and Liu, Xi and Kumar, P. R.},
  booktitle={Learning for Dynamics and Control},
  pages={815--827},
  year={2021},
  organization={PMLR}
}

@inproceedings{liu2020exploration,
  title={Exploration through reward biasing: Reward-biased maximum likelihood estimation for stochastic multi-armed bandits},
  author={Liu, Xi and Hsieh, Ping-Chun and Hung, Yu Heng and Bhattacharya, Anirban and Kumar, P. R.},
  booktitle={International Conference on Machine Learning},
  pages={6248--6258},
  year={2020},
  organization={PMLR}
}

@inproceedings{hung2021reward,
  title={Reward-biased maximum likelihood estimation for linear stochastic bandits},
  author={Hung, Yu-Heng and Hsieh, Ping-Chun and Liu, Xi and Kumar, P.R.},
  booktitle={Proceedings of the AAAI Conference on Artificial Intelligence},
  volume={35},
  pages={7874--7882},
  year={2021}
}

@article{mete2022augmented,
  title={Augmented {RBMLE-UCB} approach for adaptive control of linear quadratic systems},
  author={Mete, Akshay and Singh, Rahul and Kumar, P. R.},
  journal={Advances in Neural Information Processing Systems},
  volume={35},
  pages={9302--9314},
  year={2022}
}

@article{hung2023value,
  title={Value-Biased Maximum Likelihood Estimation for Model-based Reinforcement Learning in Discounted Linear {MDP}s},
  author={Hung, Yu-Heng and Hsieh, Ping-Chun and Mete, Akshay and Kumar, P. R.},
  journal={arXiv preprint arXiv:2310.11515},
  year={2023}
}

@inproceedings{mete2022rbmle,
  title={The {RBMLE} method for reinforcement learning},
  author={Mete, Akshay and Singh, Rahul and Kumar, P. R.},
  booktitle={2022 56th Annual Conference on Information Sciences and Systems (CISS)},
  pages={107--112},
  year={2022},
  organization={IEEE}
}

@article{schrittwieser2020mastering,
  title={Mastering {A}tari, {G}o, {C}hess and {S}hogi by planning with a learned model},
  author={Schrittwieser, Julian and Antonoglou, Ioannis and Hubert, Thomas and Simonyan, Karen and Sifre, Laurent and Schmitt, Simon and Guez, Arthur and Lockhart, Edward and Hassabis, Demis and Graepel, Thore and others},
  journal={Nature},
  volume={588},
  number={7839},
  pages={604--609},
  year={2020},
  publisher={Nature Publishing Group UK London}
}

@article{wang2024efficientzero,
  title={Efficientzero v2: Mastering discrete and continuous control with limited data},
  author={Wang, Shengjie and Liu, Shaohuai and Ye, Weirui and You, Jiacheng and Gao, Yang},
  journal={arXiv preprint arXiv:2403.00564},
  year={2024}
}

@article{ye2021mastering,
  title={Mastering {A}tari games with limited data},
  author={Ye, Weirui and Liu, Shaohuai and Kurutach, Thanard and Abbeel, Pieter and Gao, Yang},
  journal={Advances in neural information processing systems},
  volume={34},
  pages={25476--25488},
  year={2021}
}

@article{becker1981optimal,
  title={Optimal strategies for the {N}-armed bandit problem},
  author={Becker, A and Kumar, P. R.},
  journal={Univ. Maryland. Baltimore County, Math. Res. Rep},
  pages={81--1},
  year={1981}
}

@article{auer2006logarithmic,
  title={Logarithmic online regret bounds for undiscounted reinforcement learning},
  author={Auer, Peter and Ortner, Ronald},
  journal={Advances in neural information processing systems},
  volume={19},
  year={2006}
}

@article{fruit2020improved,
  title={Improved analysis of {UCRL2} with empirical bernstein inequality},
  author={Fruit, Ronan and Pirotta, Matteo and Lazaric, Alessandro},
  journal={arXiv preprint arXiv:2007.05456},
  year={2020}
}

@inproceedings{bourel2020tightening,
  title={Tightening exploration in upper confidence reinforcement learning},
  author={Bourel, Hippolyte and Maillard, Odalric and Talebi, Mohammad Sadegh},
  booktitle={International Conference on Machine Learning},
  pages={1056--1066},
  year={2020},
  organization={PMLR}
}

@inproceedings{shyam2019model,
  title={Model-based active exploration},
  author={Shyam, Pranav and Ja{\'s}kowski, Wojciech and Gomez, Faustino},
  booktitle={International conference on machine learning},
  pages={5779--5788},
  year={2019},
  organization={PMLR}
}

@article{chua2018deep,
  title={Deep reinforcement learning in a handful of trials using probabilistic dynamics models},
  author={Chua, Kurtland and Calandra, Roberto and McAllister, Rowan and Levine, Sergey},
  journal={Advances in neural information processing systems},
  volume={31},
  year={2018}
}

@article{kurutach2018model,
  title={Model-ensemble trust-region policy optimization},
  author={Kurutach, Thanard and Clavera, Ignasi and Duan, Yan and Tamar, Aviv and Abbeel, Pieter},
  journal={arXiv preprint arXiv:1802.10592},
  year={2018}
}

@article{burda2018exploration,
  title={Exploration by random network distillation},
  author={Burda, Yuri and Edwards, Harrison and Storkey, Amos and Klimov, Oleg},
  journal={arXiv preprint arXiv:1810.12894},
  year={2018}
}

@inproceedings{pathak2017curiosity,
  title={Curiosity-driven exploration by self-supervised prediction},
  author={Pathak, Deepak and Agrawal, Pulkit and Efros, Alexei A and Darrell, Trevor},
  booktitle={International conference on machine learning},
  pages={2778--2787},
  year={2017},
  organization={PMLR}
}

@article{duncan1994almost,
  title={Almost self-optimizing strategies for the adaptive control of diffusion processes},
  author={Duncan, T. E. and Pasik-Duncan, B. and Stettner, L.},
  journal={Journal of optimization theory and applications},
  volume={81},
  number={3},
  pages={479--507},
  year={1994},
  publisher={Springer}
}

@inproceedings{campi1996optimal,
  title={Optimal adaptive control of an LQG system},
  author={Campi, MC and Kumar, P. R.},
  booktitle={Proceedings of 35th IEEE Conference on Decision and Control},
  volume={1},
  pages={349--353},
  year={1996},
  organization={IEEE}
}

@article{cen2024value,
  title={Value-incentivized preference optimization: A unified approach to online and offline {RLHF}},
  author={Cen, Shicong and Mei, Jincheng and Goshvadi, Katayoon and Dai, Hanjun and Yang, Tong and Yang, Sherry and Schuurmans, Dale and Chi, Yuejie and Dai, Bo},
  journal={arXiv preprint arXiv:2405.19320},
  year={2024}
}

@article{eysenbach2022mismatched,
  title={Mismatched no more: Joint model-policy optimization for model-based RL},
  author={Eysenbach, Benjamin and Khazatsky, Alexander and Levine, Sergey and Salakhutdinov, Russ R},
  journal={Advances in Neural Information Processing Systems},
  volume={35},
  pages={23230--23243},
  year={2022}
}

@article{sutton1999policy,
  title={Policy gradient methods for reinforcement learning with function approximation},
  author={Sutton, Richard S and McAllester, David and Singh, Satinder and Mansour, Yishay},
  journal={Advances in neural information processing systems},
  volume={12},
  year={1999}
}

@article{robine2023transformer,
  title={Transformer-based world models are happy with 100k interactions},
  author={Robine, Jan and H{\"o}ftmann, Marc and Uelwer, Tobias and Harmeling, Stefan},
  journal={arXiv preprint arXiv:2303.07109},
  year={2023}
}

@article{curi2020efficient,
  title={Efficient model-based reinforcement learning through optimistic policy search and planning},
  author={Curi, Sebastian and Berkenkamp, Felix and Krause, Andreas},
  journal={Advances in Neural Information Processing Systems},
  volume={33},
  pages={14156--14170},
  year={2020}
}

@article{sukhija2025sombrl,
  title={{SOMBRL}: Scalable and Optimistic Model-Based RL},
  author={Sukhija, Bhavya and Treven, Lenart and Sferrazza, Carmelo and D{\"o}rfler, Florian and Abbeel, Pieter and Krause, Andreas},
  journal={arXiv preprint arXiv:2511.20066},
  year={2025}
}

@inproceedings{badia2020agent57,
  title={Agent57: Outperforming the {A}tari human benchmark},
  author={Badia, Adri{\`a} Puigdom{\`e}nech and Piot, Bilal and Kapturowski, Steven and Sprechmann, Pablo and Vitvitskyi, Alex and Guo, Zhaohan Daniel and Blundell, Charles},
  booktitle={International conference on machine learning},
  pages={507--517},
  year={2020},
  organization={PMLR}
}

@inproceedings{simchowitz2020naive,
  title={Naive exploration is optimal for online lqr},
  author={Simchowitz, Max and Foster, Dylan},
  booktitle={International Conference on Machine Learning},
  pages={8937--8948},
  year={2020},
  organization={PMLR}
}

@inproceedings{jedra2022minimal,
  title={Minimal expected regret in linear quadratic control},
  author={Jedra, Yassir and Proutiere, Alexandre},
  booktitle={International Conference on Artificial Intelligence and Statistics},
  pages={10234--10321},
  year={2022},
  organization={PMLR}
}

@article{janner2019trust,
  title={When to trust your model: Model-based policy optimization},
  author={Janner, Michael and Fu, Justin and Zhang, Marvin and Levine, Sergey},
  journal={Advances in neural information processing systems},
  volume={32},
  year={2019}
}

@article{kakade2020information,
  title={Information theoretic regret bounds for online nonlinear control},
  author={Kakade, Sham and Krishnamurthy, Akshay and Lowrey, Kendall and Ohnishi, Motoya and Sun, Wen},
  journal={Advances in Neural Information Processing Systems},
  volume={33},
  pages={15312--15325},
  year={2020}}

@article{wagenmaker2023optimal,
  title={Optimal exploration for model-based rl in nonlinear systems},
  author={Wagenmaker, Andrew and Shi, Guanya and Jamieson, Kevin G},
  journal={Advances in Neural Information Processing Systems},
  volume={36},
  pages={15406--15455},
  year={2023}
}

@article{liu2023maximize,
  title={Maximize to explore: One objective function fusing estimation, planning, and exploration},
  author={Liu, Zhihan and Lu, Miao and Xiong, Wei and Zhong, Han and Hu, Hao and Zhang, Shenao and Zheng, Sirui and Yang, Zhuoran and Wang, Zhaoran},
  journal={Advances in Neural Information Processing Systems},
  volume={36},
  pages={22151--22165},
  year={2023}
}

@inproceedings{bruce2024genie,
  title={Genie: Generative interactive environments},
  author={Bruce, Jake and Dennis, Michael D and Edwards, Ashley and Parker-Holder, Jack and Shi, Yuge and Hughes, Edward and Lai, Matthew and Mavalankar, Aditi and Steigerwald, Richie and Apps, Chris and others},
  booktitle={Forty-first International Conference on Machine Learning},
  year={2024}
}

@article{assran2025v,
  title={V-{JEPA} 2: Self-supervised video models enable understanding, prediction and planning},
  author={Assran, Mido and Bardes, Adrien and Fan, David and Garrido, Quentin and Howes, Russell and Muckley, Matthew and Rizvi, Ammar and Roberts, Claire and Sinha, Koustuv and Zholus, Artem and others},
  journal={arXiv preprint arXiv:2506.09985},
  year={2025}
}

@inproceedings{mete2023reward,
  title={The Reward Biased Method: An Optimism based Approach for Reinforcement Learning},
  author={Mete, Akshay and Singh, Rahul and Kumar, P. R.},
  booktitle={2023 59th Annual Allerton Conference on Communication, Control, and Computing (Allerton)},
  pages={1--7},
  year={2023},
  organization={IEEE}
}
\bibliographystyle{icml2026}

%%%%%%%%%%%%%%%%%%%%%%%%%%%%%%%%%%%%%%%%%%%%%%%%%%%%%%%%%%%%%%%%%%%%%%%%%%%%%%%
%%%%%%%%%%%%%%%%%%%%%%%%%%%%%%%%%%%%%%%%%%%%%%%%%%%%%%%%%%%%%%%%%%%%%%%%%%%%%%%
% APPENDIX
%%%%%%%%%%%%%%%%%%%%%%%%%%%%%%%%%%%%%%%%%%%%%%%%%%%%%%%%%%%%%%%%%%%%%%%%%%%%%%%
%%%%%%%%%%%%%%%%%%%%%%%%%%%%%%%%%%%%%%%%%%%%%%%%%%%%%%%%%%%%%%%%%%%%%%%%%%%%%%%
\newpage
\appendix
\onecolumn

\section{Empirical Results}\label{app:exp}
We discuss detailed results and the experimental setup in this section.   
\subsection{Hyperparameters}
The OWM framework introduces two additional hyperparameters on top of the standard world model framework, namely the optimistic bias term $\alpha(t)$ and the model entropy coefficient $\eta$. For O-DreamerV3, we set $\alpha(t)=\alpha=0.0001$ and $\eta=3 \times 10^{-6}$. For O-STORM, we set $\alpha(t)=\alpha=0.0001$ and $\eta=0.0003$. The sensitivity analysis of these parameters is provided in Figures \ref{fig:alpha_sensi} and \ref{fig:eta_sensi}.  All the other involved hyperparameters are set to the same values as the baseline algorithms and can be found in \cite{hafner2023mastering} for DreamerV3 and \cite{zhang2023storm} for STORM. 

\subsection{Results}
\subsubsection{Evaluation Methodology}  The reported learning curves are obtained using 20 evaluation episodes at each checkpoint. We plot the mean reward $\pm$ the standard error of the mean with a smoothing window of 3. The results for O-DreamerV3 on the Atari100K and the DMC benchmarks are averaged over 10 seeds. All other results, including O-STORM experiments and ablations, are averaged over 5 seeds due to computational resource limitations.

\subsubsection{Atari100K Benchmark}
\begin{figure}[ht!]
\centering
\includegraphics[scale=0.92]{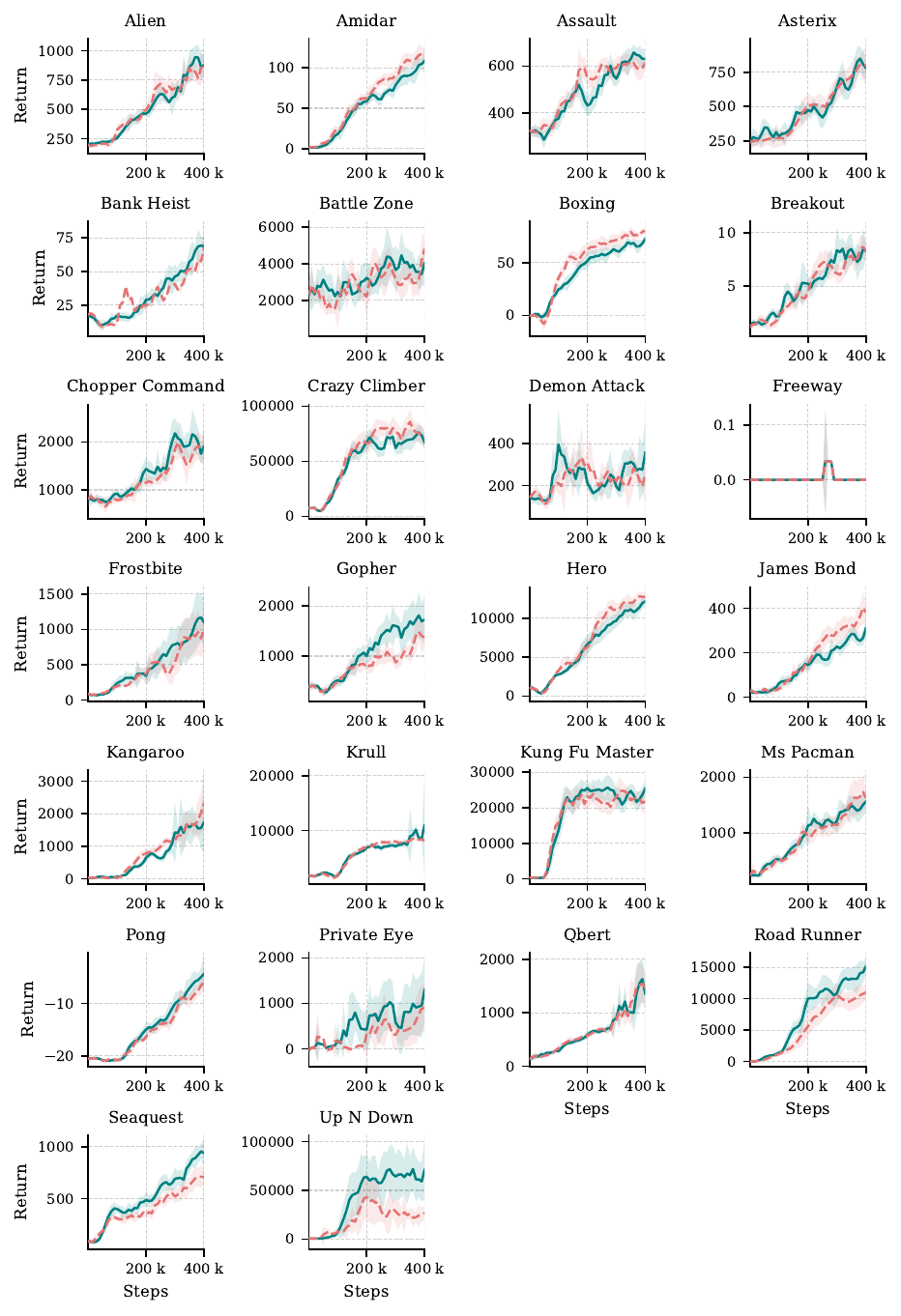}
\vspace{-1.5ex}
\includegraphics[width=0.7\linewidth]{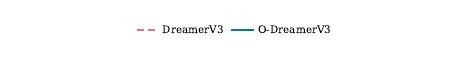}
\caption{Comparison of Optimistic DreamerV3 and DreamerV3 on the Atari100K Benchmark.}
\label{fig:atari_plots_dreamer}
\end{figure}

\begin{figure}[ht!]
\centering
\includegraphics[scale=0.92]{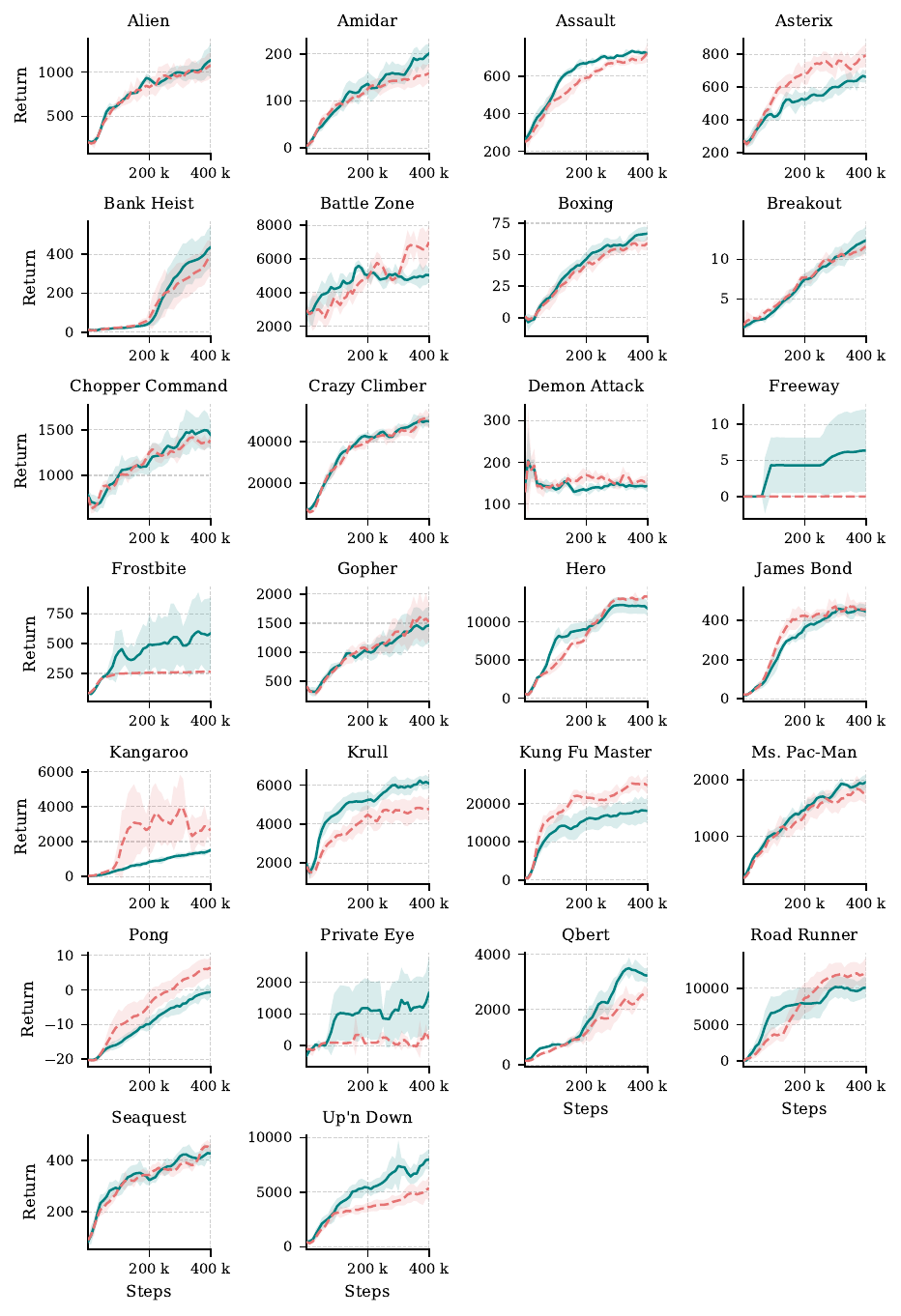}
\vspace{-1.5ex}
\includegraphics[width=0.7\linewidth]{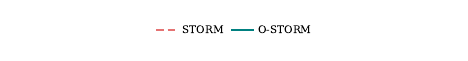}
\caption{Comparison of Optimistic STORM and STORM on the Atari100K Benchmark.}
\label{fig:atari_plots_storm}
\end{figure}

The Atari100K benchmark is designed to evaluate the data efficiency of RL algorithms. It consists of 26 Atari games, where each agent has a fixed budget of 400K frames (equivalent to roughly 2 hours of gameplay) \citep{hafner2023mastering}. The results for O-DreamerV3 are provided in Figure \ref{fig:atari_plots_dreamer} and Table \ref{tab:atari-dreamer-stats-final}. The results for O-STORM are provided in Figure \ref{fig:atari_plots_storm} and Table \ref{tab:atari100k-storm-rg}. 
\begin{table*}
\caption{Comparison of DreamerV3 and Optimistic DreamerV3 on Atari100K benchmark. Higher scores are highlighted in color and formatted in bold.}
\label{tab:atari-dreamer-stats-final}
\vskip 0.15in
\begin{center}
\begin{tabular}{lrrrrrrrr}
\toprule
& & & \multicolumn{3}{c}{DreamerV3} & \multicolumn{3}{c}{O-DreamerV3} \\
\cmidrule(lr){4-6} \cmidrule(lr){7-9}
Game & Random & Human & Mean & IQM & Median & Mean & IQM & Median \\
\midrule
Alien           & 228   & 7128  & 821                   & 800                   & 800                   & \cellcolor{green!25} \textbf{948}      & \cellcolor{green!25} \textbf{996.67}    & \cellcolor{green!25} \textbf{1065}    \\
Amidar          & 6     & 1720  & \cellcolor{red!25} \textbf{111.4} & 110.33                & 109                   & 111                          & \cellcolor{green!25} \textbf{112}       & \cellcolor{green!25} \textbf{114}     \\
Assault         & 222   & 742   & \cellcolor{red!25} \textbf{644.6} & \cellcolor{red!25} \textbf{659.67} & \cellcolor{red!25} \textbf{677}     & 621.2                        & 628.17                        & 619.5   \\
Asterix         & 210   & 8503  & 740                   & 633.33                & 550                   & \cellcolor{green!25} \textbf{805}      & \cellcolor{green!25} \textbf{858.33}    & \cellcolor{green!25} \textbf{900}     \\
Bank Heist      & 14    & 753   & \cellcolor{red!25} \textbf{79}    & \cellcolor{red!25} \textbf{73.33}  & 65                    & 67                           & 65                            & 65      \\
Battle Zone     & 2360  & 37188 & \cellcolor{red!25} \textbf{5000}  & \cellcolor{red!25} \textbf{4833.33} & \cellcolor{red!25} \textbf{4500}    & 4700                         & 3666.67                       & 3000    \\
Boxing          & 0     & 12    & \cellcolor{red!25} \textbf{79.3}  & \cellcolor{red!25} \textbf{79}     & \cellcolor{red!25} \textbf{79}      & 74.3                         & 76.67                         & 78.5    \\
Breakout        & 2     & 30    & \cellcolor{red!25} \textbf{7.9}   & \cellcolor{red!25} \textbf{7.67}   & \cellcolor{red!25} \textbf{8}       & 7.2                          & 6.83                          & 6.5     \\
Chopper Command & 811   & 7388  & 2030                  & 2050                  & \cellcolor{red!25} \textbf{2050}    & \cellcolor{green!25} \textbf{2220}     & \cellcolor{green!25} \textbf{2116.67}   & 1750    \\
Crazy Climber   & 10780 & 35829 & \cellcolor{red!25} \textbf{67390} & \cellcolor{red!25} \textbf{68616.67} & \cellcolor{red!25} \textbf{64450}   & 64190                        & 62900                         & 56700   \\
Demon Attack    & 152   & 1971  & 366                   & 193.33                & \cellcolor{red!25} \textbf{197.5}   & \cellcolor{green!25} \textbf{407.5}    & \cellcolor{green!25} \textbf{250.83}    & 180     \\
Freeway         & 0     & 30    & 0                     & 0                     & 0                     & 0                            & 0                             & 0       \\
Frostbite       & 65    & 4335  & \cellcolor{red!25} \textbf{1117}  & \cellcolor{red!25} \textbf{860}    & 300                   & 1102                         & 763.33                        & \cellcolor{green!25} \textbf{440}     \\
Gopher          & 258   & 2413  & 1356                  & 1293.33               & 1190                  & \cellcolor{green!25} \textbf{1792}     & \cellcolor{green!25} \textbf{1513.33}   & \cellcolor{green!25} \textbf{1640}    \\
Hero            & 1027  & 30826 & \cellcolor{red!25} \textbf{12799} & \cellcolor{red!25} \textbf{13368.33} & \cellcolor{red!25} \textbf{13382.5} & 12186                        & 13277.5                       & 13320   \\
James Bond      & 29    & 303   & 330                   & 325                   & 325                   & \cellcolor{green!25} \textbf{360}      & \cellcolor{green!25} \textbf{333.33}    & \cellcolor{green!25} \textbf{350}     \\
Kangaroo        & 52    & 3035  & \cellcolor{red!25} \textbf{2640}  & \cellcolor{red!25} \textbf{1566.67} & \cellcolor{red!25} \textbf{1600}    & 2140                         & 1366.67                       & 1300    \\
Krull           & 1598  & 2666  & 8320                  & \cellcolor{red!25} \textbf{8736.67} & \cellcolor{red!25} \textbf{8955}    & \cellcolor{green!25} \textbf{16816}    & 7213.33                       & 7215    \\
Kung Fu Master  & 256   & 22736 & 22890                 & 21100                 & 20850                 & \cellcolor{green!25} \textbf{26640}    & \cellcolor{green!25} \textbf{26183.33}  & \cellcolor{green!25} \textbf{26100}   \\
Ms Pacman       & 307   & 6952  & 1446                  & 1335                  & 1245                  & \cellcolor{green!25} \textbf{1552}     & \cellcolor{green!25} \textbf{1605}      & \cellcolor{green!25} \textbf{1555}    \\
Pong            & -21   & 15    & -5.6                  & -6.5                  & -7.5                  & \cellcolor{green!25} \textbf{-2.2}     & \cellcolor{green!25} \textbf{-4}        & \cellcolor{green!25} \textbf{-4.5}    \\
Private Eye     & 25    & 69571 & 893.9                 & 147.17                & 100                   & \cellcolor{green!25} \textbf{1676.4}   & \cellcolor{green!25} \textbf{1297.83}   & 100     \\
Qbert           & 164   & 13455 & \cellcolor{red!25} \textbf{1357.5} & \cellcolor{red!25} \textbf{1116.67} & \cellcolor{red!25} \textbf{1262.5}  & 1137.5                       & 787.5                         & 787.5   \\
Road Runner     & 12    & 7845  & 11330                 & 11166.67              & 11550                 & \cellcolor{green!25} \textbf{15930}    & \cellcolor{green!25} \textbf{15350}     & \cellcolor{green!25} \textbf{16050}   \\
Seaquest        & 68    & 42055 & 708                   & 660                   & 680                   & \cellcolor{green!25} \textbf{914}      & \cellcolor{green!25} \textbf{860}       & \cellcolor{green!25} \textbf{860}     \\
Up N Down       & 533   & 11693 & 24954                 & 14106.67              & 14185                 & \cellcolor{green!25} \textbf{91717}    & \cellcolor{green!25} \textbf{69470}     & \cellcolor{green!25} \textbf{51325}   \\

\bottomrule
\end{tabular}
\end{center}
\vskip -0.1in
\end{table*}

\begin{table*}
\caption{Comparison of STORM and O-STORM on the Atari 100K benchmark. Higher scores are highlighted in color and formatted in bold.}
\label{tab:atari100k-storm-rg}
\vskip 0.15in
\begin{center}
\begin{tabular}{lrrrrrrrr}
\toprule
& & & \multicolumn{3}{c}{STORM} & \multicolumn{3}{c}{O-STORM} \\
\cmidrule(lr){4-6} \cmidrule(lr){7-9}
Game & Random & Human & Mean & IQM & Median & Mean & IQM & Median \\
\midrule
Alien           & 228   & 7128  & 1100.7                     & \cellcolor{red!25} \textbf{1060.83} & \cellcolor{red!25} \textbf{1140.0}  & \cellcolor{green!25} \textbf{1131.2} & 992.5                       & 1036.5  \\
Amidar          & 6     & 1720  & 164.55                     & 153.23                      & 132.25                      & \cellcolor{green!25} \textbf{210.95} & \cellcolor{green!25} \textbf{226.9}   & \cellcolor{green!25} \textbf{231.05} \\
Assault         & 222   & 742   & \cellcolor{red!25} \textbf{749.52}  & 701.58                      & 695.35                      & 714.42                     & \cellcolor{green!25} \textbf{717.47}  & \cellcolor{green!25} \textbf{722.6}  \\
Asterix         & 210   & 8503  & \cellcolor{red!25} \textbf{785.0}   & \cellcolor{red!25} \textbf{820.0}   & \cellcolor{red!25} \textbf{840.0}   & 634.5                      & 605.83                      & 587.5   \\
Bank Heist      & 14    & 753   & 378.3                      & 376.0                       & 437.0                       & \cellcolor{green!25} \textbf{445.7}  & \cellcolor{green!25} \textbf{458.83}  & \cellcolor{green!25} \textbf{528.0}  \\
Battle Zone     & 2360  & 37188 & \cellcolor{red!25} \textbf{7470.0}  & \cellcolor{red!25} \textbf{7716.67} & \cellcolor{red!25} \textbf{8250.0}  & 5140.0                     & 4883.33                     & 5000.0  \\
Boxing          & 0     & 12    & 58.4                       & 62.3                        & 60.5                        & \cellcolor{green!25} \textbf{66.25}  & \cellcolor{green!25} \textbf{66.62}   & \cellcolor{green!25} \textbf{65.05}  \\
Breakout        & 2     & 30    & 12.23                      & \cellcolor{red!25} \textbf{12.13}  & 11.8                        & \cellcolor{green!25} \textbf{12.64}  & 11.83                       & \cellcolor{green!25} \textbf{12.25} \\
Chopper Command & 811   & 7388  & 1371.0                     & \cellcolor{red!25} \textbf{1375.0}  & 1355.0                      & \cellcolor{green!25} \textbf{1418.0} & 1358.33                     & \cellcolor{green!25} \textbf{1410.0} \\
Crazy Climber   & 10780 & 35829 & 50567.0                    & \cellcolor{red!25} \textbf{49108.33} & \cellcolor{red!25} \textbf{50710.0} & \cellcolor{green!25} \textbf{51354.0} & 48655.0                     & 48700.0 \\
Demon Attack    & 152   & 1971  & \cellcolor{red!25} \textbf{159.7}   & \cellcolor{red!25} \textbf{144.17}  & \cellcolor{red!25} \textbf{148.0}   & 148.35                     & 133.92                      & 130.5   \\
Freeway         & 0     & 30    & 0.0                        & 0.0                         & 0.0                         & \cellcolor{green!25} \textbf{6.38}   & 0.0                         & 0.0     \\
Frostbite       & 65    & 4335  & 261.5                      & \cellcolor{red!25} \textbf{261.0}   & \cellcolor{red!25} \textbf{264.5}   & \cellcolor{green!25} \textbf{586.9}  & 259.67                      & 262.0   \\
Gopher          & 258   & 2413  & \cellcolor{red!25} \textbf{1600.6}  & \cellcolor{red!25} \textbf{1416.67} & \cellcolor{red!25} \textbf{1339.0}  & 1335.2                     & 1162.67                     & 883.0   \\
Hero            & 1027  & 30826 & \cellcolor{red!25} \textbf{13215.6} & \cellcolor{red!25} \textbf{13436.83} & \cellcolor{red!25} \textbf{13433.25} & 11305.6                    & 11794.67                    & 11398.0 \\
James Bond      & 29    & 303   & \cellcolor{red!25} \textbf{460.0}   & 426.67                      & 422.5                       & 447.5                      & \cellcolor{green!25} \textbf{444.17}  & \cellcolor{green!25} \textbf{425.0}  \\
Kangaroo        & 52    & 3035  & \cellcolor{red!25} \textbf{2287.0}  & \cellcolor{red!25} \textbf{2030.0}  & 1370.0                      & 1578.0                     & 1636.67                     & \cellcolor{green!25} \textbf{1590.0} \\
Krull           & 1598  & 2666  & 4882.5                     & 5006.83                     & 5701.5                      & \cellcolor{green!25} \textbf{6294.5}  & \cellcolor{green!25} \textbf{6259.17} & \cellcolor{green!25} \textbf{6451.5} \\
Kung Fu Master  & 256   & 22736 & \cellcolor{red!25} \textbf{24545.0} & \cellcolor{red!25} \textbf{25591.67} & \cellcolor{red!25} \textbf{24025.0} & 17608.0                    & 17090.0                     & 13655.0 \\
Ms Pacman       & 307   & 6952  & 1923.0                     & \cellcolor{red!25} \textbf{2045.17} & \cellcolor{red!25} \textbf{2026.0}  & \cellcolor{green!25} \textbf{2008.5} & 1985.5                      & 1883.0  \\
Pong            & -21   & 15    & \cellcolor{red!25} \textbf{7.07}    & \cellcolor{red!25} \textbf{9.37}    & \cellcolor{red!25} \textbf{9.6}     & -0.75                      & -0.87                       & -0.65   \\
Private Eye     & 25    & 69571 & 115.53                     & 115.33                      & 100.0                       & \cellcolor{green!25} \textbf{2120.48} & \cellcolor{green!25} \textbf{1668.87} & 100.0   \\
Qbert           & 164   & 13455 & 2489.25                    & 2574.58                     & 2502.5                      & \cellcolor{green!25} \textbf{3271.0}  & \cellcolor{green!25} \textbf{3432.92} & \cellcolor{green!25} \textbf{3591.25} \\
Road Runner     & 12    & 7845  & \cellcolor{red!25} \textbf{13057.0} & \cellcolor{red!25} \textbf{12866.67} & \cellcolor{red!25} \textbf{11165.0} & 10161.0                    & 9013.33                     & 8625.0  \\
Seaquest        & 68    & 42055 & \cellcolor{red!25} \textbf{444.2}   & \cellcolor{red!25} \textbf{452.33}  & \cellcolor{red!25} \textbf{451.0}   & 421.8                      & 411.0                       & 375.0   \\
Up N Down       & 533   & 11693 & 5585.2                     & 4550.67                     & 4529.5                      & \cellcolor{green!25} \textbf{8111.0}  & \cellcolor{green!25} \textbf{7772.5}  & \cellcolor{green!25} \textbf{7888.0}  \\
\bottomrule
\end{tabular}
\end{center}
\end{table*}
\subsubsection{Deepmind Control (DMC) Suite}
\begin{figure}[ht!]
\centering
\includegraphics[width=\textwidth]{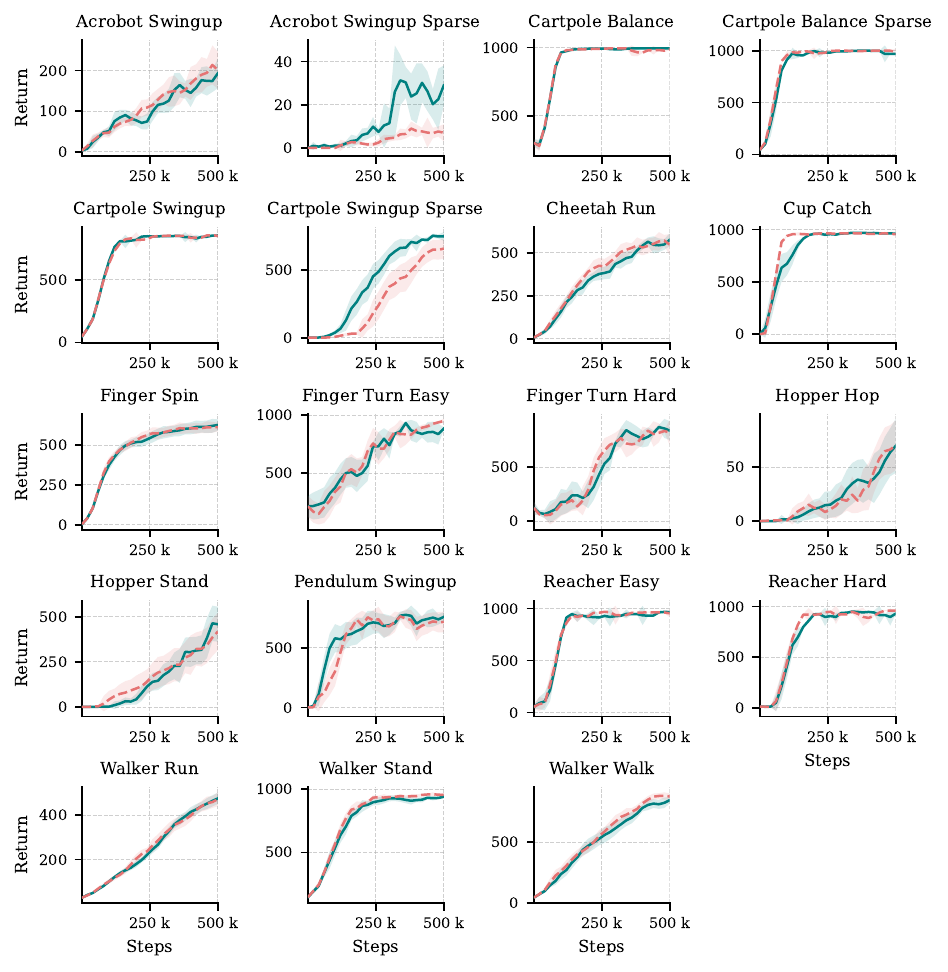}
\vspace{-1.5ex}
\includegraphics[width=0.7\linewidth]{final_plots/dreamer_legends.pdf}
\caption{Comparison of Optimistic DreamerV3 and DreamerV3 on the DMC Proprio Benchmark.}
\label{fig:atari_grid_dmc_proprio}
\end{figure}

\begin{figure}[ht!]

\centering
\includegraphics[width=\textwidth]{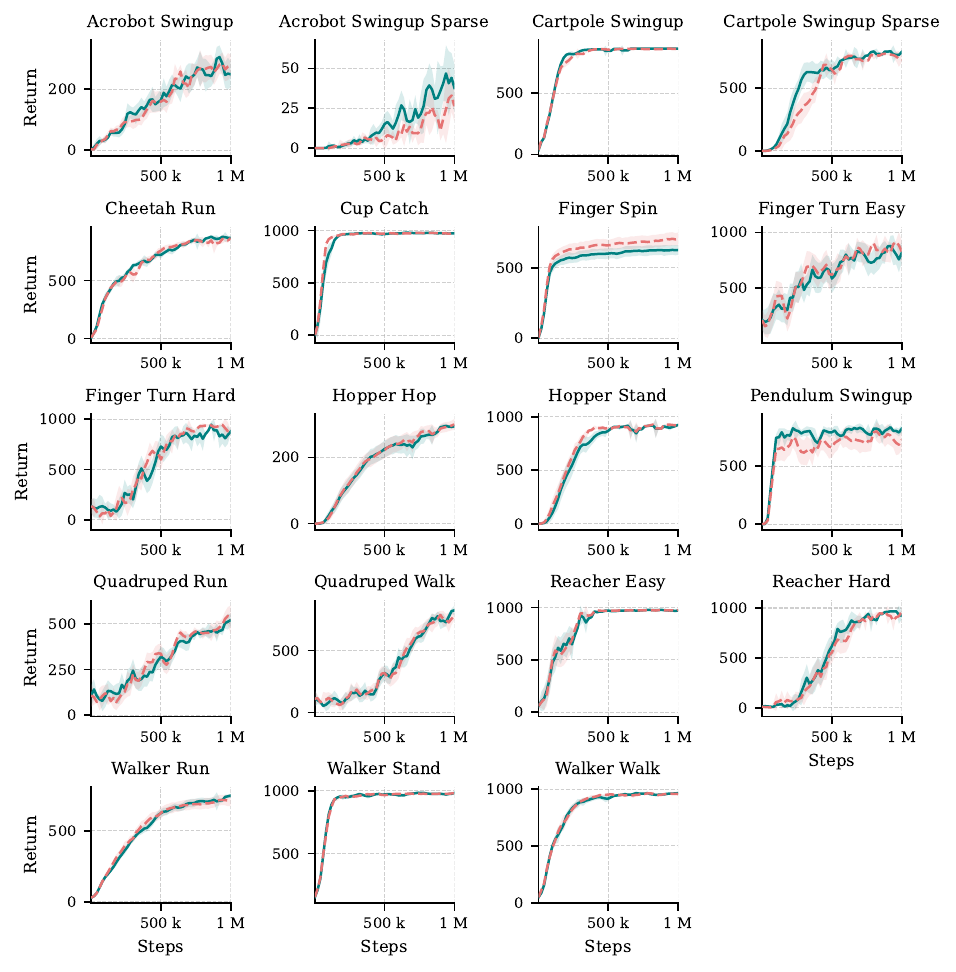}
\vspace{-1.5ex}
\includegraphics[width=0.7\linewidth]{final_plots/dreamer_legends.pdf}
\caption{Comparison of Optimistic DreamerV3 and DreamerV3 on the DMC Vision Benchmark.}
\label{fig:atari_grid_dmc_vision}
\end{figure}

We also present the performance of O-DreamerV3 on the DeepMind Control (DMC) suite, a popular reinforcement learning benchmark for continuous control tasks. The DMC suite consists of two benchmarks, the DMC proprio control benchmark and the DMC vision control benchmark. The DMC proprio control benchmark consists of continuous control tasks with proprioceptive vector inputs and continuous actions. Each task has a budget of 500K environment steps.
Meanwhile, the DMC vision control benchmark consists of continuous control tasks with high-dimensional images as inputs. Each task has a budget of 1M environment steps. The learning curves are provided in Figures \ref{fig:atari_grid_dmc_proprio} and \ref{fig:atari_grid_dmc_vision}, while the final performance is highlighted in Tables \ref{tab:dmc-proprio-stats-final} and \ref{tab:dmc-vision-stats-final}.
\begin{table*}[h!]
\caption{Comparison of Optimistic DreamerV3 and  DreamerV3 on DMC Proprio tasks. Higher scores are highlighted in color and formatted in bold.}
\label{tab:dmc-proprio-stats-final}
\vskip 0.15in
\begin{center}
\begin{tabular}{lrrrrrr}
\toprule
& \multicolumn{3}{c}{DreamerV3} & \multicolumn{3}{c}{O-DreamerV3} \\
\cmidrule(lr){2-4} \cmidrule(lr){5-7}
Task & Mean & IQM & Median & Mean & IQM & Median \\
\midrule
Acrobot Swingup         & 150.33                & 133.51                & 127.65                & \cellcolor{green!25} \textbf{236.82} & \cellcolor{green!25} \textbf{232.42} & \cellcolor{green!25} \textbf{244.35} \\
Acrobot Swingup Sparse & 8.4                   & 3.83                  & 1.5                   & \cellcolor{green!25} \textbf{34.6}   & \cellcolor{green!25} \textbf{32.5}   & \cellcolor{green!25} \textbf{28.0}   \\
Cartpole Balance         & 990.23                & \cellcolor{red!25} \textbf{995.43} & \cellcolor{red!25} \textbf{995.69} & \cellcolor{green!25} \textbf{993.0}  & 994.19                     & 995.13                     \\
Cartpole Balance Sparse& 964.0                 & 1000.0                & 1000.0                & \cellcolor{green!25} \textbf{1000.0} & 1000.0                     & 1000.0                     \\
Cartpole Swingup         & 838.83                & \cellcolor{red!25} \textbf{857.02} & \cellcolor{red!25} \textbf{857.77} & \cellcolor{green!25} \textbf{855.11} & 855.16                     & 855.81                     \\
Cartpole Swingup sparse& 664.2                 & 755.83                & 762.0                 & \cellcolor{green!25} \textbf{747.1}  & \cellcolor{green!25} \textbf{780.17} & \cellcolor{green!25} \textbf{784.0}  \\
Cheetah Run              & 522.19                & 559.83                & \cellcolor{red!25} \textbf{617.94} & \cellcolor{green!25} \textbf{591.27} & \cellcolor{green!25} \textbf{611.16} & 612.4                      \\
Cup Catch                & \cellcolor{red!25} \textbf{959.2}   & \cellcolor{red!25} \textbf{962.33} & \cellcolor{red!25} \textbf{962.0}  & 948.5                      & 957.17                     & 959.5                      \\
Finger Spin              & 609.9                 & 600.67                & 569.5                 & \cellcolor{green!25} \textbf{624.3}  & \cellcolor{green!25} \textbf{613.33} & \cellcolor{green!25} \textbf{604.5}  \\
Finger Turn Easy       & 946.9                 & 957.83                & 974.5                 & \cellcolor{green!25} \textbf{964.6}  & \cellcolor{green!25} \textbf{973.33} & \cellcolor{green!25} \textbf{980.5}  \\
Finger Turn Hard       & 703.5                 & 847.5                 & \cellcolor{red!25} \textbf{913.0}  & \cellcolor{green!25} \textbf{823.1}  & \cellcolor{green!25} \textbf{909.83} & 905.5                      \\
Hopper Hop               & \cellcolor{red!25} \textbf{86.92}   & \cellcolor{red!25} \textbf{76.14}  & \cellcolor{red!25} \textbf{62.78}  & 71.42                      & 55.92                      & 40.63                      \\
Hopper Stand             & \cellcolor{red!25} \textbf{465.78}  & \cellcolor{red!25} \textbf{481.12} & \cellcolor{red!25} \textbf{487.12} & 431.97                     & 429.69                     & 480.77                     \\
Pendulum Swingup         & 765.4                 & 843.0                 & 855.0                 & \cellcolor{green!25} \textbf{827.0}  & \cellcolor{green!25} \textbf{853.5}  & \cellcolor{green!25} \textbf{869.5}  \\
Reacher Easy           & 929.2                 & \cellcolor{red!25} \textbf{970.17} & \cellcolor{red!25} \textbf{968.5}  & \cellcolor{green!25} \textbf{965.5}  & 965.5                      & 966.5                      \\
Reacher Hard           & \cellcolor{red!25} \textbf{959.5}   & 958.0                 & 955.5                 & 959.3                      & \cellcolor{green!25} \textbf{959.33} & \cellcolor{green!25} \textbf{959.0}  \\
Walker Run               & \cellcolor{red!25} \textbf{492.96}  & \cellcolor{red!25} \textbf{498.63} & \cellcolor{red!25} \textbf{497.83} & 481.94                     & 491.25                     & 491.48                     \\
Walker Stand             & 952.7                 & \cellcolor{red!25} \textbf{969.36} & \cellcolor{red!25} \textbf{976.3}  & \cellcolor{green!25} \textbf{959.76} & 963.81                     & 967.57                     \\
Walker Walk              & 875.84                & 882.97                & 883.94                & \cellcolor{green!25} \textbf{878.62} & \cellcolor{green!25} \textbf{888.17} & \cellcolor{green!25} \textbf{903.99} \\
\bottomrule
\end{tabular}
\end{center}
\vskip -0.1in
\end{table*}
\begin{table*}
\caption{Comparison of Optimistic DreamerV3 and  DreamerV3 on DMC Vision tasks. Higher scores are highlighted in color and formatted in bold.}
\label{tab:dmc-vision-stats-final}
\vskip 0.15in
\begin{center}
\begin{tabular}{lrrrrrr}
\toprule
& \multicolumn{3}{c}{DreamerV3} & \multicolumn{3}{c}{O-DreamerV3} \\
\cmidrule(lr){2-4} \cmidrule(lr){5-7}
Task & Mean & IQM & Median & Mean & IQM & Median \\
\midrule
Acrobot Swingup         & \cellcolor{red!25} \textbf{258.64} & \cellcolor{red!25} \textbf{254.44} & 241.71                & 219.42                  & 219.97                  & \cellcolor{green!25} \textbf{249.51} \\
Acrobot Swingup Sparse& 14.2                  & 1.83                  & 0.0                   & \cellcolor{green!25} \textbf{38.6}    & \cellcolor{green!25} \textbf{28.67}   & \cellcolor{green!25} \textbf{36.0}   \\
Cartpole Swingup        & \cellcolor{red!25} \textbf{866.97} & 866.42                & 866.54                & 866.94                  & \cellcolor{green!25} \textbf{866.68}  & \cellcolor{green!25} \textbf{866.72} \\
Cartpole Swingup Sparse & 751.7                 & 758.33                & 757.0                 & \cellcolor{green!25} \textbf{795.5}   & \cellcolor{green!25} \textbf{794.17}  & \cellcolor{green!25} \textbf{791.0}  \\
Cheetah Run             & \cellcolor{red!25} \textbf{900.88} & \cellcolor{red!25} \textbf{900.95} & \cellcolor{red!25} \textbf{900.22} & 892.85                  & 897.01                  & 899.91                     \\
Cup Catch               & 971.9                 & 970.0                 & 970.0                 & \cellcolor{green!25} \textbf{973.0}   & \cellcolor{green!25} \textbf{970.33}  & 970.0                      \\
Finger Spin             & \cellcolor{red!25} \textbf{709.6}  & \cellcolor{red!25} \textbf{672.83} & 620.5                 & 629.1                   & 623.0                   & \cellcolor{green!25} \textbf{624.0}  \\
Finger Turn Easy      & 796.0                 & 937.33                & 940.0                 & \cellcolor{green!25} \textbf{849.8}   & \cellcolor{green!25} \textbf{941.83}  & \cellcolor{green!25} \textbf{946.5}  \\
Finger Turn Hard      & \cellcolor{red!25} \textbf{943.3}  & 941.67                & 943.5                 & 936.6                   & \cellcolor{green!25} \textbf{946.5}   & \cellcolor{green!25} \textbf{955.0}  \\
Hopper Hop              & \cellcolor{red!25} \textbf{303.49} & 292.11                & 289.86                & 299.31                  & \cellcolor{green!25} \textbf{298.52}  & \cellcolor{green!25} \textbf{298.87} \\
Hopper Stand            & \cellcolor{red!25} \textbf{935.54} & 930.68                & 929.79                & 932.77                  & \cellcolor{green!25} \textbf{936.21}  & \cellcolor{green!25} \textbf{933.34} \\
Pendulum Swingup        & 777.3                 & \cellcolor{red!25} \textbf{850.83} & 843.5                 & \cellcolor{green!25} \textbf{838.9}   & 844.67                  & \cellcolor{green!25} \textbf{848.5}  \\
Quadruped Run           & 482.32                & 493.46                & 481.41                & \cellcolor{green!25} \textbf{528.24}  & \cellcolor{green!25} \textbf{510.29}  & \cellcolor{green!25} \textbf{501.88} \\
Quadruped Walk          & \cellcolor{red!25} \textbf{831.16} & \cellcolor{red!25} \textbf{856.04} & \cellcolor{red!25} \textbf{867.86} & 799.58                  & 819.48                  & 840.62                     \\
Reacher Easy            & 983.2                 & 984.0                 & 984.0                 & \cellcolor{green!25} \textbf{984.6}   & 984.0                   & \cellcolor{green!25} \textbf{985.0}  \\
Reacher Hard            & \cellcolor{red!25} \textbf{961.4}  & 964.67                & 962.5                 & 933.0                   & \cellcolor{green!25} \textbf{970.17}  & \cellcolor{green!25} \textbf{972.5}  \\
Walker Run              & 714.8                 & 751.72                & 751.36                & \cellcolor{green!25} \textbf{750.3}   & \cellcolor{green!25} \textbf{756.55}  & \cellcolor{green!25} \textbf{754.9}  \\
Walker Stand            & 983.39                & 982.98                & \cellcolor{red!25} \textbf{984.18} & \cellcolor{green!25} \textbf{985.11}  & \cellcolor{green!25} \textbf{984.88}  & 983.69                     \\
Walker Walk             & 960.23                & 963.05                & 963.67                & \cellcolor{green!25} \textbf{966.63}  & \cellcolor{green!25} \textbf{968.27}  & \cellcolor{green!25} \textbf{968.07} \\
\bottomrule
\end{tabular}
\end{center}

\end{table*}
\subsection{Ablation Studies}
\begin{figure*}[t!]
  \centering
  % First row --------------------------------------------------------------
  \begin{subfigure}[b]{0.3\linewidth}
      \includegraphics[width=\linewidth]{final_plots/fig3_c_csp.pdf}
      \caption{Cartpole Swing. Sparse}
      \label{fig:10a}
  \end{subfigure}\hfill
  \begin{subfigure}[b]{0.3\linewidth}
      \includegraphics[width=\linewidth]{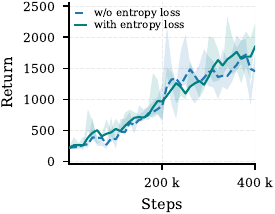}
      \caption{MsPacman}
      \label{fig:10b}
  \end{subfigure} \hfill
  \begin{subfigure}[b]{0.3\linewidth}
      \includegraphics[width=\linewidth]{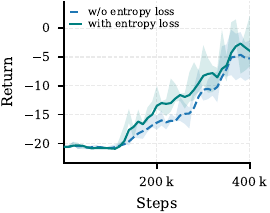}
      \caption{Pong}
      \label{fig:10c}
  \end{subfigure}  
  \caption{Ablation study of model entropy loss term for Optimistic DreamerV3}
  \label{fig:eta_ablations}
\end{figure*}

\begin{figure*}[t
!]
  \centering
  % First row --------------------------------------------------------------
  \begin{subfigure}[b]{0.3\linewidth}
      \includegraphics[width=\linewidth]{final_plots/fig3_a_csp.pdf}
      \caption{Cartpole Swing. Sparse}
      \label{fig:11a}
  \end{subfigure}\hfill
  \begin{subfigure}[b]{0.3\linewidth}
      \includegraphics[width=\linewidth]{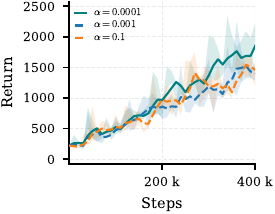}
      \caption{MsPacman}
      \label{fig:11b}
  \end{subfigure} \hfill
  \begin{subfigure}[b]{0.3\linewidth}
      \includegraphics[width=\linewidth]{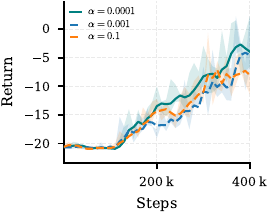}
      \caption{Pong}
      \label{fig:11c}
  \end{subfigure}  

  \caption{Performance of O-DreamerV3 for various values of $\alpha$ with $\eta=3 \times 10^{-6}$.}
  \label{fig:alpha_sensi}
\end{figure*}

\begin{figure*}[t!]
  \centering
  % First row --------------------------------------------------------------
  \begin{subfigure}[b]{0.3\linewidth}
      \includegraphics[width=\linewidth]{final_plots/fig3_b_csp.pdf}
      \caption{Cartpole Swing. Sparse}
      \label{fig:12a}
  \end{subfigure}\hfill
  \begin{subfigure}[b]{0.3\linewidth}
      \includegraphics[width=\linewidth]{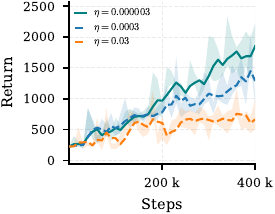}
      \caption{MsPacman}
      \label{fig:12b}
  \end{subfigure} \hfill
  \begin{subfigure}[b]{0.3\linewidth}
      \includegraphics[width=\linewidth]{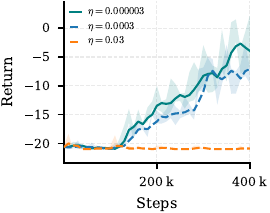}
      \caption{Pong}
      \label{fig:12c}
  \end{subfigure}  

  \caption{Performance of O-DreamerV3 for various values of $\eta$ with $\alpha=0.0001$.}
  \label{fig:eta_sensi}
\end{figure*}

Our optimistic dynamics loss consists of two terms, the optimism term and the entropy term.
Figure \ref{fig:eta_ablations} shows the performance without the entropy loss term, highlighting the improvement due to the introduction of the entropy loss. 
The sensitivity analysis of the two hyperparameters $\alpha$ and $\eta$ is provided in Figures \ref{fig:alpha_sensi} and \ref{fig:eta_sensi} respectively. These results show that high values, such as $\alpha=0.1$, $\eta=0.03$ are detrimental to performance, while smaller values enhance performance. 
\subsection{Computation Resources}
The O-DreamerV3 code is implemented using the official open-source code of DreamerV3 available at \url{https://github.com/danijar/DreamerV3}. Similarly, O-STORM is implemented using the official STORM code available at \url{https://github.com/weipu-zhang/STORM}. Each experiment was completed using one A100 GPU. The computational time experiments were run on a local RTX 4090 GPU to ensure fairness.

\end{document}